\documentclass[a4paper,12pt]{article}
\usepackage[super,square]{natbib}
\usepackage{anysize}
\marginsize{2cm}{2cm}{1cm}{2cm}
\usepackage{fancyhdr}

\usepackage{soul}
\usepackage[dvipsnames]{xcolor}
\usepackage{hyperref}
\hypersetup{
    colorlinks=true,
    linkcolor=black,
    citecolor=CadetBlue,
    filecolor=CadetBlue,      
    urlcolor=CadetBlue,
}
\usepackage{lipsum}
\usepackage{multicol}
\usepackage{parskip}
\renewenvironment{abstract}
 {\par\noindent\textbf{\abstractname}\ \ignorespaces \\}
 {\par\noindent\medskip}

\usepackage{amsmath} 
\usepackage{tabularx}
\usepackage{booktabs}
\usepackage{float} 
\usepackage{graphicx} 
\usepackage{amssymb}
\usepackage{ragged2e}

\begin{document}
\pagestyle{fancy}
\thispagestyle{empty}
\fancyhead[R]{\textit{}}
\fancyhead[L]{}
\renewcommand*{\thefootnote}{\fnsymbol{footnote}}
\begin{center}
\Large{\textbf{A comprehensive taxonomy of hallucinations in Large Language Models}}
\vspace{0.4cm}
\normalsize
\\ Manuel Cossio MMed, MEng \\
\vspace{0.1cm}
\textit{Universitat de Barcelona} \\
\vspace{0.1cm}
\textit{manuel.cossio@ub.edu}
\medskip
\normalsize
\end{center}
{\color{gray}\hrule}
\vspace{0.4cm}
\begin{abstract}
Large language models (LLMs) have revolutionized natural language processing, yet their propensity for "hallucination"—generating plausible but factually incorrect or fabricated content—remains a critical challenge. This report provides a comprehensive taxonomy of LLM hallucinations, beginning with a formal definition and a theoretical framework that posits its inherent inevitability in computable LLMs, irrespective of architecture or training. It explores core distinctions, differentiating between intrinsic (contradicting input context) and extrinsic (inconsistent with training data or reality), as well as factuality (absolute correctness) and faithfulness (adherence to input). The report then details specific manifestations, including factual errors, contextual and logical inconsistencies, temporal disorientation, ethical violations, and task-specific hallucinations across domains like code generation and multimodal applications. It analyzes the underlying causes, categorizing them into data-related issues, model-related factors, and prompt-related influences. Furthermore, the report examines cognitive and human factors influencing hallucination perception, surveys evaluation benchmarks and metrics for detection, and outlines architectural and systemic mitigation strategies. Finally, it introduces web-based resources for monitoring LLM releases and performance. This report underscores the complex, multifaceted nature of LLM hallucinations and emphasizes that, given their theoretical inevitability, future efforts must focus on robust detection, mitigation, and continuous human oversight for responsible and reliable deployment in critical applications.
\end{abstract}
{\color{gray}\hrule}
\medskip

\tableofcontents

\section{Introduction}

Large language models (LLMs) represent a significant advancement in natural language processing (NLP), fundamentally altering how information is acquired and processed\cite{huang2025survey, elchafei2025span}. These models have enabled a paradigm shift, facilitating diverse applications ranging from sophisticated content creation to advanced decision support systems \cite{huang2025survey, elchafei2025span}. Their capacity to generate human-like text has led to remarkable progress in various tasks, including complex question-answering systems, abstractive summarization, and interactive conversational agents\cite{roustan2025clinicians}. The widespread adoption of LLMs underscores their transformative potential across numerous industries and research domains.

Despite their impressive capabilities, a critical and widely acknowledged limitation of LLMs is their propensity for "hallucination" \cite{huang2025survey, elchafei2025span, joshi2025mitigating, lee2025hallucination}. This phenomenon describes the generation of content that, while often plausible and coherent, is factually incorrect, inconsistent, or entirely fabricated \cite{huang2025survey, elchafei2025span}. Unlike the medical definition of hallucination, which refers to sensory experiences in the absence of external stimuli, in the context of LLMs, it signifies the creation of nonfactual information to respond to a user's query, frequently without any explicit indication of its fabricated nature \cite{ohayon2000prevalence}. Such generated content is characterized as incorrect, nonsensical, and lacking justifiable basis, making its detection challenging for users. The prevalence of hallucinations raises significant concerns regarding the reliability and trustworthiness of LLMs, particularly as their integration into real-world information retrieval (IR) systems and critical decision-making processes continues to expand \cite{huang2025survey}.

This report provides a comprehensive taxonomy of hallucinations in Large Language Models (LLMs), delving into the critical challenge of their propensity for generating plausible but factually incorrect or fabricated content. It begins by formally defining hallucination, including a framework that posits its inherent inevitability in computable LLMs, irrespective of their architecture or training. The core distinctions of hallucination are explored, differentiating between intrinsic (contradicting input context) and extrinsic (inconsistent with training data or reality), as well as factuality (absolute correctness) and faithfulness (adherence to input) \cite{huang2025survey}.

The report then details specific manifestations of hallucinations, such as factual errors, contextual and logical inconsistencies, temporal disorientation, ethical violations, and task-specific hallucinations in domains like code generation and multimodal applications \cite{tian2025codehalu}. A thorough analysis of the underlying causes is presented, categorizing them into data-related issues (e.g., quality, biases), model-related factors (e.g., auto-regressive nature, decoding strategies, lack of reasoning), and prompt-related influences (e.g., adversarial attacks) \cite{huang2025survey, joshi2025mitigating}.

The report further examines the cognitive and human factors influencing hallucination perception, including user trust, interpretability, and various cognitive biases, alongside their implications for design and mitigation strategies \cite{yoo2024much}. It provides a comprehensive overview of evaluation benchmarks and metrics for hallucination detection, surveying principal datasets and quantitative metrics, as well as their current limitations. Additionally, a detailed exploration of hallucination mitigation strategies is included, covering both architectural approaches (e.g., Toolformer-style augmentation, factual grounding through retrieval mechanisms) and systemic approaches (e.g., guardrails, symbolic integration) \cite{schick2023toolformer, joshi2025mitigating}. Finally, the report introduces crucial web-based resources for monitoring LLM releases and performance, offering insights into tools and leaderboards that track hallucination rates and model reliability in real-world scenarios. This report underscores the complex, multifaceted nature of LLM hallucinations and emphasizes that, given their theoretical inevitability, future efforts must focus on robust detection, mitigation, and continuous human oversight for responsible and reliable deployment in critical applications.

\vspace{0.5cm}
{\color{gray}\hrule}
\begin{center}
\section{Defining hallucination in LLMs}
\textbf{This section formally defines LLM hallucination and presents a theoretical framework arguing for its inherent inevitability.}
\bigskip
\end{center}
{\color{gray}\hrule}

\subsection{General conceptual definition}

In the domain of LLMs, hallucination is broadly understood as the generation of "plausible yet nonfactual content"\cite{huang2025survey}. This implies that an LLM produces "false or fabricated information" or outputs that are "inaccurate, irrelevant, or simply does not make factual sense" \cite{joshi2025mitigating, lee2025hallucination}. A key distinction from the medical definition of hallucination (sensory experiences without corresponding external stimuli) is that in LLMs, it refers to the creation of nonfactual content in response to a user's question, often without the model clarifying the fabricated nature of its answer\cite{huang2025survey}. This characteristic underscores the challenge of relying on LLM outputs without external verification.

\subsection{Formal definition and inevitability}

The paper "Hallucination is Inevitable: An Innate Limitation of Large Language Models" offers a formal framework for understanding hallucination, defining it within a "formal world" of computable functions to rigorously analyze its inherent inevitability in LLMs\cite{xu2024hallucination}.

\subsubsection{Formal definition}

Hallucination is formally defined as an inconsistency between a computable LLM, denoted as $h$, and a computable ground truth function, $f$.
\begin{itemize}
    \item \textbf{Formal world of $f$ (ground truth function):} this is conceptualized as a set $G_f = \{(s, f(s)) | s \in S\}$, where $f(s)$ represents the \textit{sole} correct output for any given input string $s$ from the set of all finite-length strings $S$\cite{xu2024hallucination}.
    \item \textbf{Training samples $T$:} these are defined as a collection of input-output pairs \\ $\{(s_0, y_0), (s_1, y_1),... | y_i = f(s_i)\}$ derived from the formal world of $f$. This set $T$ serves as a generalized corpus representing the expected outputs of $f$ for corresponding inputs\cite{xu2024hallucination}.
    \item \textbf{Hallucination condition:} an LLM $h$ is considered to be "hallucinating" with respect to a ground truth function $f$ if, across \textit{all} training stages $i$ (meaning, after being trained on any finite number of samples), there \textit{exists} at least one input string $s$ for which the LLM's output $h[i](s)$ does not match the correct output $f(s)$\cite{xu2024hallucination}. This condition is formally expressed as $\forall i \in \mathbb{N}, \exists s \in S \text{ such that } h[i](s) \neq f(s)$.
\end{itemize}

\subsubsection{Implications for inevitability}

The paper posits that hallucination is an inevitable characteristic of LLMs, irrespective of their architectural design, learning algorithms, prompting techniques, or the specific training data employed, provided they are considered "computable LLMs" operating within the defined formal world\cite{xu2024hallucination}. The central argument supporting this claim is rooted in diagonalization, a proof technique used in computability theory to demonstrate that certain infinite sets are inherently larger than others, implying limitations on what can be computed.

This theoretical framework leads to several critical theorems:

\begin{itemize}
    \item \textbf{Theorem 1: computably enumerable LLMs will hallucinate:} this theorem states that for any computably enumerable set of LLMs (a category that includes all currently proposed polynomial-time bounded LLMs), there exists a computable ground truth function $f$ such that \textit{all} states of \textit{all} LLMs within that set will exhibit hallucination\cite{xu2024hallucination}. This is demonstrated by constructing a ground truth function $f$ that is specifically designed to contradict the output of every LLM state along a diagonal enumeration of all LLM states and their outputs.
    \item \textbf{Theorem 2: LLMs will hallucinate on infinitely many questions:} building upon the first theorem, this extends the argument to assert that for any computably enumerable set of LLMs, there exists a computable ground truth function $f$ such that \textit{all} states of \textit{all} LLMs in that set will hallucinate on an \textit{infinite number} of inputs\cite{xu2024hallucination}. This is shown by constructing an $f$ that consistently differs from the output of each LLM state for an unending sequence of inputs.
    \item \textbf{Theorem 3: any computable LLM will hallucinate:} this theorem generalizes the preceding findings. It asserts that for \textit{any individual computable LLM}, there exists a computable ground truth function $f$ such that \textit{every state} of that LLM will hallucinate with respect to $f$. Furthermore, for any computable LLM, there exists another computable ground truth function $f'$ for which every state of that LLM will hallucinate on \textit{infinitely many} inputs\cite{xu2024hallucination}. This theorem holds particular significance because real-world LLMs are considered a subset of total computable LLMs, directly extending the theoretical inevitability to practical applications.
    \item \textbf{Corollary 1: inability to self-eliminate hallucination:} a direct consequence of Theorem 3 is that \textit{all computable LLMs inherently lack the capacity to prevent themselves from hallucinating}\cite{xu2024hallucination}. This implies that mitigation strategies relying solely on the LLM's internal mechanisms, such as prompt-based chain-of-thought reasoning, cannot fully eliminate hallucination.
\end{itemize}

\textbf{Practical implications of inevitability:} the formal definition and the theorems supporting the inevitability of hallucination carry profound practical implications for the development and deployment of LLMs:
\begin{itemize}
    \item \textbf{General problem solvers:} LLMs, when trained solely on input-output pairs and employed as general problem solvers, are inherently prone to hallucination, particularly for problems that are computationally hard or lie beyond their learned capabilities\cite{xu2024hallucination}.
    \item \textbf{Scrutiny of mathematical and logic reasoning:} outputs from LLMs concerning mathematical problems and logic reasoning should always be subjected to rigorous scrutiny, as these domains often involve computationally challenging tasks that increase the likelihood of hallucination\cite{xu2024hallucination}.
    \item \textbf{Safety-critical decisions:} without the integration of external aids such as guardrails, knowledge bases, or direct human control, LLMs cannot be autonomously used in safety-critical decision-making processes. Human oversight remains paramount for decisions demanding rational and humane judgment\cite{xu2024hallucination}.
    \item \textbf{Research and regulations:} the inherent inevitability of hallucination underscores the urgent need for rigorous study and the establishment of appropriate regulations concerning the safety boundaries of LLMs. This is crucial to ensure their sustainable development and prevent their deployment in contexts that exceed their inherent capabilities\cite{xu2024hallucination}.
\end{itemize}

This observation highlights a fundamental aspect of hallucination: it is not merely a "bug" or an "error" that can be entirely eradicated through improved training or architectural design. Instead, it is an innate limitation rooted in the very nature of computability. If hallucination is formally proven to be inevitable for \textit{any computable LLM}, it fundamentally redefines the objective from complete elimination to robust reduction and effective management. This understanding necessitates a paradigm shift in how LLMs are conceptualized, evaluated, and deployed. Rather than striving for perfect factual accuracy, the focus must pivot towards designing systems that incorporate robust detection mechanisms, containment strategies, and, crucially, human-in-the-loop validation, especially for applications where accuracy is paramount. This foundational understanding reinforces the indispensable role of external aids, such as Retrieval-Augmented Generation (RAG) systems, and direct human intervention, as LLMs cannot fully self-correct this inherent limitation (see summary in Table \ref{tab:llm_theorems}).

\begin{table}[H]
\centering
\caption{Theorems and corollaries on LLM hallucination}
\vspace{0.3cm}
\label{tab:llm_theorems}
\renewcommand{\arraystretch}{1.3}
\setlength{\tabcolsep}{6pt}
\small
\begin{tabularx}{\textwidth}{>{\bfseries\raggedright\arraybackslash}X>{\raggedright\arraybackslash}X>{\raggedright\arraybackslash}X>{\raggedright\arraybackslash}X}
\toprule
\textbf{Theorem/corollary} & \textbf{Statement} & \textbf{Implication for real-world LLMs} & \textbf{Reference} \\
\midrule

Theorem 1: computably enumerable LLMs will hallucinate & 
For any computably enumerable set of LLMs, there exists a computable ground truth function f such that all states of all LLMs in that set will hallucinate. & 
All currently proposed polynomial-time bounded LLMs are inherently prone to hallucination; it cannot be completely eliminated. & \cite{xu2024hallucination}\\
\addlinespace[3pt]

Theorem 2: LLMs will hallucinate on infinitely many questions & 
For any computably enumerable set of LLMs, there exists a computable ground truth function f such that all states of all LLMs in that set will hallucinate on infinitely many inputs. & 
Hallucinations are not isolated incidents but a persistent challenge across a vast range of inputs for any LLM. & \cite{xu2024hallucination}\\
\addlinespace[3pt]

Theorem 3: any computable LLM will hallucinate & 
For any individual computable LLM, there exists a computable ground truth function f such that every state of that LLM will hallucinate. Furthermore, for any computable LLM, there exists another f' such that every state will hallucinate on infinitely many inputs. & 
This generalizes inevitability to any specific LLM, confirming that current and future LLMs will always exhibit some form of hallucination. & \cite{xu2024hallucination}\\
\addlinespace[3pt]

Corollary 1: inability to self-eliminate hallucination & 
All computable LLMs cannot prevent themselves from hallucinating. & 
LLMs cannot solely rely on internal mechanisms (e.g., self-correction, chain-of-thought prompting) to eliminate hallucination; external safeguards are essential. & \cite{xu2024hallucination}\\
\bottomrule
\end{tabularx}
\end{table}

\vspace{0.5cm}
{\color{gray}\hrule}
\begin{center}
\section{Core taxonomies of LLM hallucinations}
\textbf{This section outlines the primary categorizations of LLM hallucinations, distinguishing between intrinsic vs. extrinsic and factuality vs. faithfulness.}
\bigskip
\end{center}
{\color{gray}\hrule}
The scientific literature presents several key categorizations for LLM hallucinations, reflecting different perspectives on their nature and origin. Two widely accepted and fundamental distinctions are between intrinsic vs. extrinsic hallucinations and factuality vs. faithfulness hallucinations (see summary in Table \ref{tab:taxonomy}).

\subsection{ Intrinsic vs. extrinsic hallucinations}

This dichotomy is a widely accepted and foundational distinction within the taxonomy of LLM hallucinations.\cite{ elchafei2025span, orgad2024llms, wang2024videohallucer, bang2025hallulens} It differentiates errors based on their relationship to the provided input context and the model's internal knowledge.

\begin{itemize}
    \item \textbf{Intrinsic hallucination:}
    intrinsic hallucinations refer to generated text that directly contradicts the \textit{provided input or context}\cite{bang2025hallulens, orgad2024llms}. These errors arise from logical inconsistencies \textit{within the generated output itself}, without necessarily requiring reference to external knowledge\cite{bang2025hallulens}. This type of hallucination reflects the model's inability to maintain consistency during the inference process or limitations stemming from its internal knowledge and parametric memory\cite{bang2025hallulens}. It can also encompass instances where the model misinterprets or omits crucial details from a given document, leading to an inaccurate representation of the source information.\cite{bang2025hallulens}

    For example, if an article provided for summarization states that the FDA approved the first Ebola vaccine in 2019, an intrinsic hallucination would manifest as a summary claiming that the FDA rejected it. Another illustrative instance is a model summarizing an article that states a person was born in 1980, and then, later in the same summary, incorrectly claiming they were born in 1975, thereby demonstrating an internally inconsistent response.

    \item \textbf{Extrinsic hallucination:}
    extrinsic hallucinations, conversely, refer to generated text that is \textit{not consistent with the training data} and "can neither be supported nor refuted by the input context"\cite{rafi2025reducing, bang2025hallulens}. This category involves the introduction of entities, facts, or events that do not exist in reality. Such hallucinations frequently occur when models generate novel content or attempt to bridge perceived knowledge gaps\cite{rafi2025reducing, bang2025hallulens}. This phenomenon highlights the model's limitations in fully absorbing knowledge from its training data and its inability to accurately recognize the boundaries of its own knowledge. It can also result from issues related to integrating external information or from the model misinterpreting or failing to correctly incorporate the given context or prompt\cite{rafi2025reducing, bang2025hallulens}.

    An example of an extrinsic hallucination is the claim that "The Parisian Tiger was hunted to extinction in 1885," a fabricated entity and event. Similarly, if a summarization article states that the FDA approved the first Ebola vaccine in 2019, an extrinsic hallucination might be a summary claiming that China started testing a COVID-19 vaccine, introducing information unrelated to the provided context.
\end{itemize}

\subsection{ Factuality vs. faithfulness hallucinations}

This represents another prevalent categorization of LLM hallucinations, focusing on the truthfulness of the generated content and its adherence to the input\cite{li2024dawn}.

\begin{itemize}
    \item \textbf{Factuality hallucination:}
    factuality hallucination occurs when an LLM generates "factually incorrect content".\cite{joshi2025mitigating, li2024dawn} This type of hallucination directly contradicts "real-world knowledge" or "established verification sources". It pertains to the "absolute correctness of the content generated" when compared against verifiable information. These errors often arise due to the model's limited contextual understanding and the inherent noise or inaccuracies present in its training data, leading to responses that are not grounded in reality\cite{cao2021hallucinated, li2024dawn}.

    Examples include the model claiming "Charles Lindbergh was the first to walk on the moon" , stating that "The Great Wall of China is visible from space", or generating the statement, "The speed of light in a vacuum is 100,000 km/s," when the correct value is approximately 299,792 km/s. Another instance is the assertion that "Thomas Edison invented the internet".

    \item \textbf{Faithfulness hallucination:}
    faithfulness errors occur when the model's output "diverges from the input prompt or provided context" \cite{maynez2020faithfulness, vishwanath2024faithfulness, malin2025review}. The response generated by the model may be internally consistent and appear plausible, but it fails to align with the user's expectations or the specific information explicitly provided in the input. This type of hallucination is closely related to, and often overlaps with, intrinsic hallucination, as both deal with inconsistencies relative to the given source \cite{maynez2020faithfulness, vishwanath2024faithfulness, malin2025review}.

    For example, in the context of summarization, if an article states that the FDA approved the first Ebola vaccine in 2019, a faithfulness hallucination would include a summary claiming that the FDA rejected it, directly contradicting the provided source information.
\end{itemize}

The presence of multiple, slightly different, yet often overlapping taxonomies (e.g., intrinsic/extrinsic versus factuality/faithfulness) across various scientific articles \cite{huang2025survey, bang2025hallulens, orgad2024llms, rafi2025reducing, joshi2025mitigating, li2024dawn, maynez2020faithfulness, vishwanath2024faithfulness, malin2025review} indicates that the field is still actively defining and refining the categorization of LLM hallucinations. While intrinsic and faithfulness hallucinations largely describe deviations from \textit{provided context} or \textit{internal consistency}, extrinsic and factuality hallucinations relate to inconsistencies with \textit{external knowledge} or \textit{real-world truth}. This nuance is critical because different types of hallucinations often stem from distinct underlying mechanisms and, consequently, require specific detection and mitigation strategies. For instance, Retrieval-Augmented Generation (RAG) is frequently cited as an effective method to combat factual or extrinsic hallucinations by grounding the model in external, verifiable knowledge\cite{aboulela2025exploring}. In contrast, intrinsic hallucinations might necessitate more sophisticated internal consistency checks or improvements in the model's reasoning capabilities. This observation underscores that the absence of a "unified framework due to inconsistent definitions and categorizations" is a significant challenge in benchmarking hallucinations\cite{bang2025hallulens, orgad2024llms}. This implies that comparative research on hallucination rates and the development of universally applicable mitigation strategies are hindered by the lack of standardized terminology and evaluation metrics. Future research efforts should prioritize the development of a more harmonized and widely accepted taxonomy to enable more effective and comparable evaluations across different models and tasks, ultimately accelerating progress in addressing this critical issue.

\vspace{0.5cm}

{\color{gray}\hrule}
\begin{center}
\section{Specific categories and manifestations of hallucinations}
\textbf{This section details various specific types of hallucinations, including factual errors, contextual inconsistencies, and task-specific manifestations.}
\bigskip
\end{center}
{\color{gray}\hrule}
Beyond the core intrinsic/extrinsic and factuality/faithfulness distinctions, LLM hallucinations manifest in numerous specific forms, often with distinct characteristics and implications (see summary in Table \ref{tab:taxonomy}).

\subsection{Factual errors and fabrications}

This is a prevalent and particularly dangerous type of LLM hallucination, characterized by the generation of incorrect, misleading, or entirely fabricated factual content, frequently presented with a high degree of confidence. Such errors can appear as inaccuracies in historical information, scientific facts, or biographical details\cite{chandler2024detecting}.

\subsubsection{Incorrect facts}
These are claims that directly oppose established and verified information\cite{chandler2024detecting, azaria2023internal}. An example is Google Bard's hallucination claiming the James Webb Space Telescope took the first images of an exoplanet, despite NASA's records indicating that earlier images existed. Other instances include the assertion that "The Great Wall of China is visible from space"  or the statement that "Thomas Edison invented the internet".

\subsubsection{Fabricated entities/information}

This involves the invention of historical figures, events, or specific details that do not exist in reality. This can extend to creating entirely fictitious narratives, such as a claim about "unicorns in Atlantis" being documented in 10,000 BC. In legal contexts, this type of hallucination can be particularly severe, involving the fabrication of information, including fake quotes and citations of non-existent court cases, leading to significant professional and legal consequences\cite{padiu2024extent}. Similarly, in medical contexts, models may fabricate clinical details, invent research citations, or create made-up disease details, posing substantial risks to patient care\cite{li2025detecting, byun2024reference}.

\subsubsection{Adversarial attacks}

A specific subset of factual errors arises from adversarial attacks, where deliberately or inadvertently fabricated details embedded within user prompts lead the model to produce or elaborate on false information. This phenomenon can result in a "garbage in, garbage out" problem, where erroneous inputs propagate misleading outputs, and also presents a threat of malicious misuse, where bad actors could exploit LLMs to spread falsehoods \cite{li2025detecting, xu2023llm, zou2023universal}.

\subsection{ Contextual inconsistencies}

Contextual inconsistencies occur when the model's output includes information not present in the provided context or directly contradicts it. This type of hallucination is often referred to as "context divergence"  or "contextual misalignment" , indicating the model's difficulty in correctly attending to relevant context and instead relying on its internal generative tendencies. An example is when the model is given the context: "The Nile originates in Central Africa," but responds with: "The Nile originates in the mountain ranges of Central Africa," adding incorrect details not found in the original input \cite{joshi2025mitigating, an2024make, elchafei2025span}.

\subsection{Instruction inconsistencies/deviation}

Instruction inconsistencies refer to instances where the LLM ignores or fails to follow specific instructions provided by the user. The generated response, in these cases, does not adhere to the user's explicit directives. For example, if instructed to translate a question into Spanish, the model might instead provide the answer in English\cite{yin2023llm}.

\subsection{ Logical inconsistencies}

Logical inconsistencies manifest when the model's output contains internal logical errors or contradictions, even if the initial part of the response is correct. This can appear as self-contradictory statements within the same output or across different interaction instances. This type of hallucination is related to "erroneous inference hallucination"  and accounts for a notable portion, specifically 19\%, of identified hallucination cases \cite{joshi2025mitigating, lee2025hallucination, ghosh2024logical, uceda2024reasoning}. An example is an LLM performing an arithmetic operation incorrectly within a step-by-step mathematical solution, or stating a fact in one sentence and then providing a conflicting statement later in the same response

\subsection{ Temporal disorientation}

Temporal disorientation describes a type of hallucination involving issues with time-sensitive information, leading to the generation of outdated, anachronistic, or temporally incorrect facts. LLMs are particularly noted for struggling with "intricate temporal features" and out-of-distribution knowledge related to time. This category accounts for 12\% of identified hallucination cases.\cite{lee2025hallucination, li2025detecting} An illustrative example is an LLM incorrectly asserting that "Haruki Murakami won the Nobel Prize in Literature in 2016," when in fact, he has not won the Nobel Prize.

\subsection{ Ethical violations}

Ethical violations refer to hallucinations that result in harmful, defamatory, or legally incorrect content. These instances can have severe real-world consequences, impacting individuals' reputations, causing financial losses, or leading to legal repercussions. Ethical violations represent 6\% of hallucination cases in some analyses.\cite{lee2025hallucination, jiao2024navigating, gan2024navigating}

\subsubsection{Defamation/misinformation}
Examples include ChatGPT falsely claiming a university professor made sexually suggestive comments and attempted to touch a student, citing a non-existent article \cite{joshi2025mitigating, cheong2022envisioning}. Another case involved ChatGPT falsely accusing a mayor of bribery and imprisonment, when he was actually a whistleblower.

\subsubsection{Financial misinformation}
An AI chatbot providing incorrect refund information to a customer, resulting in financial loss for both the customer and the airline, exemplifies how hallucinations can lead to tangible economic harm\cite{yoo2024much}.

\subsubsection{Legal inaccuracies}
LLMs can produce content that deviates from actual legal facts, well-established legal principles, or precedents. This includes generating "bogus judicial decisions, bogus quotes, and bogus internal citations". Such errors can lead to "representational harm," where the contributions of one member of the legal community are systematically erased or misattributed \cite{li2025detecting, cheong2024not}.

\subsection{ Amalgamated hallucinations}

Amalgamated hallucinations occur when the model incorrectly combines multiple facts or conditions presented within a single prompt. This happens when the LLM fails to properly integrate several distinct conditions, resulting in a blended output that erroneously merges disparate pieces of information\cite{elchafei2025span, zhang2024knowledge}.

\subsection{ Nonsensical responses}

Nonsensical responses are instances where LLMs generate output that is completely irrelevant to the input prompt. This type highlights the model's limitations in understanding context or maintaining a logical thread in a conversation, posing significant challenges in user interaction scenarios where clarity and relevance are paramount.\cite{joshi2025mitigating} An example is a conversation about the NBA Commissioner where the LLM initially mentions "Adam Silver" but then randomly switches to "Stern" in the same response.

\subsection{ Task-specific hallucinations}

Hallucinations can manifest uniquely depending on the specific generative task the LLM is performing.

\subsubsection{Dialogue history-based hallucination}
This occurs when an LLM mixes up names or relations of entities from the conversation history, or creates new incorrect inferences based on previous errors, leading to a "snowball effect" of distorted context. This arises because LLMs rely on pattern recognition and statistics, often lacking common sense or factual grounding in dialogue\cite{xu2024hallucination, dziri2022origin}.

\subsubsection{Abstractive summarization hallucination}

Systems designed for abstractive summarization can introduce errors or semantic transformations between the original and generated data, distorting or fabricating details, inferring unsupported causal relationships, or retrieving unrelated background knowledge. This is attributed to their reliance on pattern recognition rather than true comprehension of the source text \cite{xu2024hallucination, kryscinski2019evaluating, maynez2020faithfulness}.

\subsubsection{Generative question answering hallucination}

In this context, the LLM makes an erroneous inference from its source information, leading to an incorrect answer, even when relevant source material is provided. The model may ignore evidence and make unjustified inferences based on its own prior knowledge \cite{xu2024hallucination, snyder2024early}.

\subsubsection{Code generation hallucination}
 When generating source code, LLMs can produce incorrect, nonsensical, or unjustifiable code that is difficult to identify and fix, especially under specific execution paths. This undermines the trustworthiness of generated code and can introduce significant risks and errors into codebases. Existing surveys classify these into input-conflicting, context-conflicting, and fact-conflicting types\cite{liu2024exploring, agarwal2024codemirage}.

\subsubsection{Multimodal large language models hallucination}

In multimodal large language models (MLLMs), hallucinations primarily focus on the "discrepancy between generated text response and provided visual content," a phenomenon known as cross-modal inconsistency \cite{huang2025survey, wu2025combating}. Object hallucination in MLLMs is empirically categorized into three types:
    \begin{itemize}
        \item \textbf{Category:} identifies nonexistent or incorrect object categories in a given image \cite{huang2025survey, wu2025combating}.
        \item \textbf{Attribute:} emphasizes incorrect descriptions of objects' attributes (e.g., color, shape, material)\cite{huang2025survey, wu2025combating}.
        \item \textbf{Relation:} assesses incorrect relationships between objects\cite{huang2025survey, wu2025combating}.
    \end{itemize}

\subsection{ Complexities and critical implications of diverse hallucination types}

The extensive list of specific hallucination types (factual, contextual, logical, temporal, ethical, amalgamated, nonsensical) and their distinct manifestations across various applications (dialogue, summarization, QA, code generation, multimodal) underscores that hallucination is not a singular, uniform error. Each type often arises from different underlying mechanisms \cite{chandler2024detecting, azaria2023internal,joshi2025mitigating, lee2025hallucination, li2025detecting, cheong2024not,snyder2024early, liu2024exploring, agarwal2024codemirage, wu2025combating}. For example, temporal errors might stem from outdated data, logical inconsistencies from reasoning flaws, and ethical violations from training biases. The detailed examples, particularly from the medical \cite{ li2025detecting, byun2024reference} and legal \cite{li2025detecting, cheong2024not} domains, vividly illustrate that these are not merely academic curiosities but critical safety, reliability, and ethical issues with significant real-world repercussions, such as misleading clinicians, misinforming patients, legal sanctions, reputational damage, and financial loss. This granular understanding implies that a "one-size-fits-all" solution for hallucination is unlikely to be effective. Instead, research and development must adopt a highly granular and context-aware approach, tailoring detection, prevention, and mitigation strategies to the specific type of hallucination prevalent in a given application domain. This also highlights the urgent need for domain-specific benchmarks and evaluation frameworks to accurately assess and address these diverse forms of factual and contextual divergence.

\begin{table}[H]
\centering
\caption{Comprehensive taxonomy of LLM hallucinations}
\vspace{0.3cm}
\label{tab:taxonomy}
\renewcommand{\arraystretch}{1.3}
\setlength{\tabcolsep}{6pt}
\small
\begin{tabularx}{\textwidth}{>{\bfseries}p{0.15\textwidth}X>{\RaggedRight}p{0.35\textwidth}>{\RaggedRight}p{0.15\textwidth}}
\toprule
\textbf{Type} & \textbf{Definition/description} & \textbf{Example} & \textbf{Sources} \\
\midrule

\textbf{Intrinsic} & Contradicts provided input or context; internal inconsistencies. & Summary states birth year as 1980 then 1975. & \cite{bang2025hallulens, orgad2024llms} \\

\textbf{Extrinsic} & Not consistent with training data; introduces non-existent entities. & ``The Parisian Tiger was hunted to extinction in 1885.'' & \cite{rafi2025reducing, bang2025hallulens} \\

\textbf{Factuality} & Contradicts real-world knowledge or verification sources. & ``Charles Lindbergh was first to walk on the moon.'' & \cite{joshi2025mitigating, li2024dawn, cao2021hallucinated} \\

\textbf{Faithfulness} & Diverges from input prompt or context. & Summary claims FDA rejected vaccine when article stated approval. & \cite{maynez2020faithfulness, vishwanath2024faithfulness, malin2025review} \\

\textbf{Factual Errors} & Incorrect, misleading, or fabricated content. & Bard claiming JWST took first exoplanet images. & \cite{chandler2024detecting} \\

\textbf{Contextual} & Contradicts or adds to provided context. & Input: ``Nile in Central Africa.'' Output: ``Nile in Central African mountains.'' & \cite{joshi2025mitigating, an2024make, elchafei2025span} \\

\textbf{Instruction} & Fails to follow user instructions. & Translates question to Spanish but answers in English. & \cite{yin2023llm} \\

\textbf{Logical} & Internal logical errors or contradictions. & Incorrect arithmetic in step-by-step solution. & \cite{joshi2025mitigating, lee2025hallucination, ghosh2024logical, uceda2024reasoning} \\

\textbf{Temporal} & Time-sensitive errors and anachronisms. & ``Murakami won Nobel Prize in 2016.'' & \cite{lee2025hallucination, li2025detecting} \\

\textbf{Ethical} & Harmful, defamatory or legally incorrect content. & False accusation of professor with non-existent citation. & \cite{lee2025hallucination, jiao2024navigating, gan2024navigating} \\

\textbf{Amalgamated} & Incorrectly combines multiple facts. & (Blending disparate information) & \cite{elchafei2025span, zhang2024knowledge} \\

\textbf{Nonsensical} & Irrelevant responses lacking logic. & Switches from ``Adam Silver'' to ``Stern'' in NBA discussion. & \cite{joshi2025mitigating} \\

\textbf{Code generation} & Incorrect or nonsensical source code. & Illogical code unfaithful to requirements. & \cite{liu2024exploring, agarwal2024codemirage} \\

\textbf{Multimodal} & Text-visual content discrepancies. & Identifying non-existent object in image. & \cite{huang2025survey, wu2025combating} \\

\bottomrule
\end{tabularx}
\end{table}

\vspace{0.5cm}
{\color{gray}\hrule}
\begin{center}
\section{Underlying causes of hallucinations}
\textbf{This section explores the complex factors contributing to hallucinations, stemming from training data, model architecture, and user prompts.}
\bigskip
\end{center}
{\color{gray}\hrule}

The diverse manifestations of LLM hallucinations stem from a complex interplay of factors originating from the training data, the model's architecture and learning processes, and the nature of user prompts (see summary in Table \ref{tab:causes_LLM}) .

\subsection{ Data-related factors}

The quality and characteristics of the data used to train LLMs are fundamental determinants of hallucination frequency and type.

\subsubsection{Quality and volume of training dataset}

The inherent quality and sheer volume of the dataset upon which an LLM is trained are crucial variables directly influencing the frequency and specific types of hallucinations produced. Flawed, incomplete, or noisy training data—containing errors, inconsistencies, or irrelevant information—significantly contributes to the generation of factually incorrect responses\cite{ joshi2025mitigating, gautam975impact}.

\subsubsection{Inadequate representation and biases}

If the data used to train LLMs lacks sufficient quality or diversity, the model may struggle to accurately understand the complexities and nuances of human language. Training on incorrect or biased data can lead to "imitative falsehoods," where the model replicates misinformation present in its training corpus \cite{joshi2025mitigating, dai2024bias}.

\subsubsection{Outdated training data and knowledge boundary}

LLMs are prone to disseminating misinformation, particularly concerning frequently updated topics, primarily due to the static nature of their training data\cite{chenghaozhu2025your}. The absence of up-to-date facts leads to inherent limitations in specialized domains. A critical issue is the model's inability to recognize its own knowledge boundaries, which often results in it confidently generating content beyond its learned scope \cite{ren2023investigating}.

\subsubsection{Source-reference divergence}

In certain datasets, such as those specifically curated for summarization tasks, summaries might contain additional, unsupported claims that diverge from the original source references, directly contributing to hallucination \cite{bn2025fact}.

\subsection{ Model-related factors}

The internal design, training processes, and inference mechanisms of LLMs contribute significantly to their propensity for hallucination.

\subsubsection{Auto-regressive nature}
A fundamental cause of hallucinations arises from the very design principle of certain LLMs: their auto-regressive nature. These models are programmed to produce output based on token prediction, meaning they predict the most probable next token(s) given the preceding sequence. Factual accuracy is not the direct, explicit goal of this process; rather, accuracy is inferred from a high probability of adequate token prediction based on the training data. Since training datasets are necessarily flawed or incomplete, the probabilistic nature of this generation can lead to hallucinations\cite{li2024dawn, huang2025survey}.

\subsubsection{Architecture flaws and internal design}
The internal design of LLMs inherently predisposes them to generating hallucinations. For instance, unidirectional representation in certain architectures can limit contextual understanding, leading to the generation of one-sided or biased narratives\cite{ huang2025survey}.

\subsubsection{Training processes}

\begin{itemize}
        \item \textbf{Exposure bias:} a discrepancy between the conditions encountered during training and those during inference can cause cascading errors during text generation.\cite{pozzi2025mitigating}
        \item \textbf{Capability misalignment:} when LLMs are aligned with capabilities that extend beyond what their training data adequately supports, they may produce errors, particularly fabricated facts in specialized domains where their knowledge is insufficient \cite{lu2025maximum}.
        \item \textbf{Belief misalignment:} the generated outputs may diverge from the LLM's internal "beliefs" or learned representations, leading to inaccuracies. This can sometimes be a result of the model "pandering" to user opinions rather than adhering to factual truth \cite{kirk2023understanding}.
        \item \textbf{Over-optimization for specific objectives:} over-optimization during the training phase for certain performance metrics can inadvertently increase hallucination rates in other areas \cite{lin2023mitigating}.
    \end{itemize}

\subsubsection{Decoding strategies}

\begin{itemize}
        \item \textbf{Stochastic nature/inherent sampling randomness:} LLMs employ sampling strategies during text generation that introduce an element of randomness into the output. A high "temperature" setting, for example, can enhance creativity but also significantly elevate the risk of hallucination by favoring the selection of low-probability or unexpected tokens \cite{huang2025survey}.
        \item \textbf{Imperfect decoding representation:} issues such as over-reliance on partially generated content and the "softmax bottleneck" can lead to faithfulness errors, where the output deviates from the intended meaning or context \cite{chang2022softmax}.
    \end{itemize}

\subsubsection{Overconfidence and calibration}

LLMs frequently exhibit overconfidence, generating outputs with high certainty even when the underlying information is incorrect. Poor calibration, where confidence scores do not accurately reflect prediction accuracy, can mislead users, particularly clinicians in medical contexts, into trusting inaccurate outputs, posing significant risks \cite{chhikara2025mind}.

\subsubsection{Generalization to unseen cases}

LLMs may struggle to generalize effectively beyond their training data, especially when confronted with rare diseases, novel treatments, or atypical clinical presentations. In such scenarios, models might extrapolate from unrelated patterns, producing erroneous or irrelevant outputs\cite{lin2023mitigating, joshi2025mitigating, dai2024bias, gautam975impact}.

\subsubsection{Lack of reasoning and nuanced language understanding}

LLMs primarily rely on statistical correlations learned from vast amounts of text rather than true causal or logical reasoning. This can lead to outputs that sound plausible but lack logical coherence. Furthermore, they struggle with interpreting the subtleties of human language, including irony, sarcasm, and cultural references, which can result in outdated or irrelevant information when nuance is key. A lack of logical reasoning capabilities is specifically identified as a significant contributor to fact-conflicting hallucinations \cite{joshi2025mitigating, jin2023can}.

\subsubsection{Knowledge overshadowing}

This occurs when certain aspects of a prompt disproportionately dominate the model's attention, leading to an overgeneralization of dominant conditions or patterns. This phenomenon is partly attributable to imbalances within the training data \cite{zhang2025law}.

\subsubsection{Insufficient knowledge representation}

Hallucinations can also arise from deficiencies in the model's internal knowledge representation, particularly within the lower layers of its neural network. These deficiencies result in what are termed "knowledge enrichment hallucinations," where the model generates unsupported information due to gaps in its subject-specific knowledge\cite{feng2024don}.

\subsubsection{Failure in information extraction}

Inaccurate extraction of relevant attributes or details by the model's higher-layer attention mechanisms can lead to "answer extraction hallucinations." This underscores the critical importance of precise information retrieval and application in generating correct outputs\cite{zhang2024knowledge}.

\subsection{ Prompt-related factors}

The way users interact with LLMs through prompts can also induce or exacerbate hallucinations.

\subsubsection{Adversarial attacks}

Deliberate or inadvertent fabrications embedded in user prompts can trigger hallucinations, as LLMs may elaborate on the false information. This creates a "garbage in, garbage out" problem, where erroneous inputs produce misleading outputs, and also poses a threat of malicious misuse\cite{xu2023llm}.

\subsubsection{Overly confirmatory tendency}

Some LLMs exhibit an overly confirmatory tendency, sometimes prioritizing a persuasive or confident style over factual accuracy. This characteristic can exacerbate the impact of prompt-based fabrications, making the hallucinated content appear more credible\cite{omar2025benchmarking}.

\subsubsection{Prompting methods}

The specific methods and clarity of prompting can significantly influence hallucination rates.  Clearer, more restrictive prompts and providing relevant in-context learning examples (e.g., few-shot learning) can help reduce hallucinations by guiding the model more precisely\cite{huang2025survey}.

\subsection{An emergent property requiring systemic solutions}

The comprehensive list of causes, spanning data quality, model architecture, training processes, and inference mechanisms, reveals that hallucination is not a simple bug but an an emergent property of the current LLM design paradigm. The auto-regressive nature \cite{li2024dawn, huang2025survey} fundamentally prioritizes generating plausible token sequences based on statistical patterns rather than ensuring factual accuracy or logical coherence. This statistical reliance, combined with "black box reasoning"  and inherent "overconfidence" \cite{chhikara2025mind}, creates a scenario where models confidently produce incorrect information. The struggle with "logical reasoning" \cite{joshi2025mitigating, jin2023can} and "generalization to unseen cases" \cite{lin2023mitigating, joshi2025mitigating, dai2024bias, gautam975impact} points to a deeper limitation beyond mere memorization; LLMs currently lack true comprehension and causal understanding. Furthermore, the vulnerability to "adversarial attacks" and the "garbage in, garbage out" problem \cite{xu2023llm} highlight the fragility of these systems to external inputs. Effectively addressing hallucination, therefore, requires a multi-pronged research agenda that goes beyond superficial fixes like data cleaning or simple fine-tuning. It necessitates fundamental advancements in model architectures to incorporate stronger symbolic reasoning capabilities, better uncertainty quantification, and more robust grounding mechanisms, such as advanced retrieval-augmented generation (RAG) techniques. The theoretical inevitability of hallucination \cite{bang2025hallulens} further reinforces that some level of hallucination might always persist, making external safeguards, robust evaluation, and continuous human oversight crucial for the safe and reliable deployment of LLMs in critical applications.

\begin{table}[H]
\centering
\caption{Root Causes of LLM Hallucinations}
\vspace{0.3cm}
\label{tab:causes_LLM}
\renewcommand{\arraystretch}{1.3}
\setlength{\tabcolsep}{6pt}
\small
\begin{tabularx}{\textwidth}{>{\bfseries\RaggedRight}p{0.15\textwidth}>{\RaggedRight}p{0.2\textwidth}>{\RaggedRight}p{0.45\textwidth}>{\RaggedRight}p{0.2\textwidth}}
\toprule
\textbf{Category} & \textbf{Specific Factor} & \textbf{Explanation} & \textbf{Sources} \\
\midrule

\textbf{Data} & 
Training Data Quality & 
Flawed, incomplete, or noisy data leads to incorrect responses. & 
\cite{joshi2025mitigating, gautam975impact} \\

& 
Data Biases & 
Lack of diversity causes imitative falsehoods. & 
\cite{joshi2025mitigating, dai2024bias} \\

& 
Outdated Data & 
Static data causes misinformation on dynamic topics. & 
\cite{chenghaozhu2025your, ren2023investigating} \\

& 
Source Divergence & 
Summaries containing unsupported claims. & 
\cite{bn2025fact} \\

\textbf{Model} & 
Auto-Regressive Nature & 
Token prediction prioritizes probability over accuracy. & 
\cite{li2024dawn, huang2025survey} \\

& 
Architecture Flaws & 
Design predisposes models to hallucinate. & 
\cite{huang2025survey} \\

& 
Exposure Bias & 
Training-inference discrepancy causes errors. & 
\cite{pozzi2025mitigating} \\

& 
Capability Misalignment & 
Fabrication in specialized domains. & 
\cite{lu2025maximum} \\

& 
Belief Misalignment & 
Outputs diverge from internal representations. & 
\cite{kirk2023understanding} \\

& 
Over-optimization & 
Focus on metrics increases hallucinations. & 
\cite{lin2023mitigating} \\

& 
Sampling Randomness & 
High temperature introduces inaccuracies. & 
\cite{huang2025survey} \\

& 
Decoding Issues & 
Over-reliance on partial generation. & 
\cite{chang2022softmax} \\

& 
Overconfidence & 
High certainty for incorrect outputs. & 
\cite{chhikara2025mind} \\

& 
Generalization Failure & 
Errors on rare/novel cases. & 
\cite{lin2023mitigating, joshi2025mitigating, dai2024bias, gautam975impact} \\

& 
Reasoning Limits & 
Statistical over causal reasoning. & 
\cite{joshi2025mitigating, jin2023can} \\

& 
Knowledge Overshadowing & 
Prompt aspects dominate attention. & 
\cite{zhang2025law} \\

& 
Knowledge Gaps & 
Deficient internal representations. & 
\cite{feng2024don} \\

& 
Extraction Failure & 
Inaccurate attention mechanisms. & 
\cite{zhang2024knowledge} \\

\textbf{Prompt} & 
Adversarial Attacks & 
Fabricated prompt details. & 
\cite{xu2023llm} \\

& 
Confirmatory Bias & 
Persuasive style over facts. & 
\cite{omar2025benchmarking} \\

& 
Poor Prompting & 
Unclear structure increases errors. & 
\cite{huang2025survey} \\

\bottomrule
\end{tabularx}
\end{table}

\vspace{0.5cm}
{\color{gray}\hrule}
\begin{center}
\section{Cognitive and human factors in hallucination perception}
\textbf{This section explores how human trust, cognitive biases, and interaction design influence the perception and impact of LLM hallucinations, emphasizing the need for user-centered mitigation strategies and human-in-the-loop oversight.}
\bigskip
\end{center}
{\color{gray}\hrule}

In addition to technical causes, the real-world impact of hallucinations is strongly shaped by how humans interpret, trust, and respond to language model outputs. Research in human-computer interaction (HCI), psychology, and decision science indicates that users are not passive consumers of information—they bring cognitive biases, heuristics, and trust dynamics into their interactions with LLMs. These factors influence whether hallucinations are detected, ignored, or acted upon.

\subsection{User trust and interpretability}

LLMs often produce fluent, well-structured, and grammatically correct responses, which are commonly interpreted by users as signals of credibility—even when the content is factually incorrect. This ``\textit{fluency heuristic}'' has been observed to increase perceived accuracy of statements simply due to linguistic polish~\cite{reber1999effects}.

Moreover, large-scale studies have shown that users tend to assign high trust to AI outputs, particularly when models present information confidently or with detailed elaboration~\cite{binns2018s}. For instance, Bubeck et al.~\cite{bubeck2023sparks} found that users often rated GPT-4’s incorrect answers as more convincing than correct ones from human experts in blind evaluations.

\subsection{Hallucinations often go unnoticed}

Because hallucinations are often contextually plausible and stylistically convincing, users—especially non-experts—may struggle to identify falsehoods without access to external verification tools. This risk is particularly high in areas like health, law, or finance, where subtle distortions can have serious consequences~\cite{li2025detecting, cheong2024not}.

Empirical studies by Luger \& Sellen~\cite{luger2016like} reveal that users often accept AI-generated outputs at face value and fail to notice hallucinations—particularly when responses appear fluent and confident—unless they are explicitly instructed to fact-check, indicating a widespread tendency to overtrust AI systems as reliable or authoritative sources.

\subsection{Cognitive biases amplifying hallucination risks}

Several well-established cognitive biases contribute to the tendency to overlook or accept hallucinated content.

\subsubsection{Automation bias}

This refers to the human tendency to over-rely on automated systems, assuming their outputs are accurate—even when they are not. In the context of LLMs, users may accept incorrect or hallucinated information simply because it comes from an AI, especially in situations involving time pressure, cognitive overload, or lack of expertise. This bias can lead users to overlook obvious errors or fail to cross-check information they would otherwise question~\cite{cummings2017automation}.

\subsubsection{Confirmation bias}

This bias describes the tendency to favor information that confirms one's pre-existing beliefs or expectations, while dismissing or overlooking contradictory evidence. When interacting with LLMs, users may be more likely to accept hallucinated content if it aligns with what they already believe or want to be true, making them less likely to scrutinize its accuracy~\cite{nickerson1998confirmation, o2025confirmation}.

\subsubsection{Illusion of explanatory depth}

This cognitive bias occurs when individuals believe they understand complex topics more deeply than they actually do. As a result, they may overestimate their ability to evaluate the accuracy of AI-generated content. When an LLM produces a plausible-sounding explanation or summary, users may assume it is correct without fully understanding or verifying the underlying concepts, increasing the risk of accepting hallucinated information~\cite{rozenblit2002misunderstood, mehta2025embracing}.

\subsubsection{Persistence of biases despite warnings}

Research shows that cognitive biases such as automation bias, confirmation bias, and the illusion of explanatory depth can persist even when users are explicitly informed that AI systems may produce errors~\cite{dzindolet2003role}. In experimental settings, users who were told that a decision-support system was fallible still tended to trust its outputs over their own judgment. This suggests that merely warning users about possible inaccuracies is often insufficient to prevent overreliance. In the context of LLMs, this means that even transparent disclaimers or uncertainty indicators may not fully mitigate the undue influence of confident but hallucinated outputs, especially when users lack the domain expertise or motivation to verify them independently.

\subsection{Design implications and mitigation strategies}

These findings suggest that hallucination mitigation is not solely a model-centric challenge, but also a user interface and interaction design problem. To improve user resilience against hallucinations, several strategies have been proposed.

\subsubsection{Calibrated uncertainty displays}

Providing users with visual or textual indicators of a model’s confidence—such as probability scores, uncertainty ranges, or qualitative labels (e.g., ``highly confident,'' ``low certainty'')—can help them better judge the reliability of AI outputs~\cite{lin2022teaching}. These displays are especially valuable in tasks like question answering or medical advice, where the perceived confidence of a model often influences user trust. When confidence is misaligned with correctness (e.g., high confidence in a hallucinated answer), users may be misled unless the interface communicates epistemic uncertainty clearly. Calibrated uncertainty helps users decide when additional verification is necessary and supports a more cautious interpretation of potentially hallucinated content.

\subsubsection{Source-grounding indicators}

 Clearly linking parts of the model’s output to supporting evidence from external sources—such as documents retrieved through a RAG system—can reduce blind acceptance of hallucinated facts~\cite{zeng2024justilm}. Visual markers, citations, or tooltips that explain which parts of a response are grounded in specific documents enhance transparency and user understanding. By making the boundary between supported and unsupported content explicit, source-grounding indicators help users identify which claims are verifiable and which may be speculative or invented, thus mitigating the impact of hallucinations in high-stakes applications.

\subsubsection{Justification prompts}

Designing systems that encourage users to ask reflective questions like ``Why is this the answer?'' or ``How do you know that?'' promotes more critical evaluation of LLM responses~\cite{favero2024enhancing}. These prompts can be implemented through interface design (e.g., buttons or suggested queries) or integrated into conversational flows. Encouraging justification-seeking behavior not only increases user awareness of potential inaccuracies but also reinforces an epistemic mindset in which outputs are evaluated based on evidence and reasoning rather than surface plausibility. This can be especially helpful in educational or decision-support contexts where understanding the rationale behind a response is as important as the response itself.

\subsubsection{Factuality-aware interface prototypes}

 Recent research has produced interface prototypes—such as Med-PaLM 2~\cite{singhal2023large}—that integrate design features aimed at improving interpretability and factual reliability. For instance, Med-PaLM 2 provides clinical references and confidence levels in its medical responses, demonstrating how multimodal transparency cues—combining visual, textual, and interactive elements—can enhance user awareness of potential hallucinations, promote responsible usage, and support informed decision-making, particularly in high-stakes domains like healthcare and public safety.

\subsection{Human-in-the-loop evaluation and oversight}

Ultimately, hallucination detection and management must be seen as a joint cognitive task between the LLM and its user. Evaluation frameworks should therefore include human factors—such as susceptibility to bias, trust calibration, and verification behavior—as part of their assessment.

This aligns with broader calls in AI safety and responsible AI literature for systems that are not just high-performing in benchmarks, but usable, trustworthy, and robust under real-world conditions~\cite{holstein2019improving, doshi2017towards}.

\vspace{0.5cm}
{\color{gray}\hrule}
\begin{center}
\section{Evaluation benchmarks and metrics for hallucination detection}
\textbf{This section surveys the principal benchmarks and evaluation metrics developed to detect and quantify hallucinations in LLMs, highlighting current methodologies, their limitations, and the need for unified, taxonomy-aware assessment frameworks.}
\bigskip
\end{center}
{\color{gray}\hrule}

The effective detection and quantification of hallucinations in LLMs is a prerequisite for both empirical research and practical deployment. While considerable progress has been made in identifying hallucination types and underlying causes, the evaluation of hallucination remains a challenging and evolving area. This section presents a brief survey of the most prominent benchmarks and metrics used to assess hallucination in LLM outputs, alongside a discussion of their limitations and future directions.

\subsection{Benchmark datasets}

Several benchmarks have been developed to systematically evaluate hallucinations across diverse tasks and domains. These datasets vary in scope, annotation methodology, and underlying definition of hallucination. \cite{lin2021truthfulqa, bang2025hallulens, min2023factscore, honovich2021q, scialom2021questeval}

\subsubsection{TruthfulQA}
 Is a benchmark composed of adversarially constructed questions that intentionally target common misconceptions, false beliefs, or ambiguities in general knowledge. Unlike traditional benchmarks that evaluate correctness in terms of expected facts, TruthfulQA emphasizes a model’s robustness against confidently generating plausible-sounding but factually incorrect statements. It is task-agnostic and domain-general, designed to test whether language models can distinguish between fact and fiction in open-domain question answering. The benchmark includes both multiple-choice and free-form response settings and is annotated with human-verified judgments to assess truthfulness, informativeness, and consistency. \cite{lin2021truthfulqa}

\subsubsection{HalluLens}
 Is a comprehensive benchmark that systematically maps hallucination instances to an explicit taxonomy encompassing multiple dimensions: factual, ethical, logical, temporal, and task-specific hallucinations. Unlike task-bound benchmarks, HalluLens is designed to evaluate hallucinations across a wide range of contexts and generation types, including summarization, question answering, dialogue, and instruction following. Each instance is annotated with detailed metadata specifying the hallucination category, severity, and grounding status. This makes it particularly suitable for fine-grained, taxonomy-aware evaluation and enables rigorous cross-model comparison aligned with theoretical frameworks. HalluLens serves as both a diagnostic and comparative tool, helping researchers and developers identify model-specific hallucination patterns. \cite{bang2025hallulens} 

\subsubsection{FActScore}

FActScore is a benchmark specifically designed to evaluate the factual consistency of outputs in summarization tasks. Rather than relying on surface-level similarity metrics such as ROUGE or BLEU, FActScore employs entailment-based classifiers that have been fine-tuned to determine whether a generated sentence can be logically inferred from the corresponding reference source document. This allows it to detect subtle hallucinations, such as fabricated relationships or omitted qualifiers, which might not be flagged by traditional overlap-based metrics. By providing span-level and sentence-level assessments, FActScore supports a more granular and semantically precise evaluation of summary fidelity. \cite{min2023factscore} 

\subsubsection{Q2 and QuestEval}

These metrics adopt an indirect yet powerful approach to evaluating factual consistency through question generation and answering. \textbf{Q2} (Quality Questioning) generates a set of questions based on the system output and then uses the source document to answer them. If the answers from the source align with those implied by the generated summary, the output is considered faithful.  \cite{honovich2021q}  \textbf{QuestEval}, similarly, computes consistency by comparing answers to questions generated from both the candidate and reference texts. These methods do not rely on static reference texts but instead treat the source content as a dynamic knowledge base, allowing for flexible and contextual evaluation of hallucination. Their strength lies in capturing factual divergences that traditional string-matching metrics often overlook. \cite{scialom2021questeval} 

\subsubsection{Domain-specific benchmarks}

Benchmarks developed to evaluate hallucinations in specialized and high-risk applications such as medicine, software engineering, and multimodal reasoning. \cite{pandit2025medhallu, zuo2024medhallbench, pal2023med, tian2025codehalu, seth2024hallucinogen}

\begin{itemize}
\item \textbf{MedHallu} this is a comprehensive benchmark specifically designed for detecting medical hallucinations in LLMs. It comprises 10,000 high-quality question-answer pairs derived from PubMedQA, with hallucinated answers systematically generated through a controlled pipeline. \cite{pandit2025medhallu} 

\item \textbf{MedHallBench} another recent and comprehensive benchmark framework for evaluating and mitigating hallucinations in Medical LLMs. It integrates expert-validated medical case scenarios with established medical databases. \cite{zuo2024medhallbench} 

\item \textbf{Med-HALT} this benchmark proposes a two-tiered approach to evaluate the presence and impact of hallucinations in biomedical-generated LLM outputs. It includes Reasoning Hallucination Tests (RHTs) like False Confidence Test (FCT), None of the Above (Nota) Test, and Fake Questions Test (FQT), as well as Memory Hallucination Tests (MHTs) such as Abstract-to-Link and PMID-to-Title tests. \cite{pal2023med}

\item \textbf{CodeHaluEval} targets the evaluation of hallucinations in code-generating language models (CodeLLMs). It includes programming tasks and ground truth outputs annotated for syntactic validity, semantic correctness, and adherence to functional requirements. The benchmark categorizes hallucinations into input-conflicting, context-conflicting, and fact-conflicting errors. It plays a crucial role in identifying code that may appear plausible but fails to execute correctly or violates specifications—risks that can lead to software bugs, security vulnerabilities, or production failures. \cite{tian2025codehalu} 

\item \textbf{HALLUCINOGEN benchmark}  is a novel and comprehensive Visual Question Answering (VQA) benchmark specifically designed to evaluate and identify "object hallucination" in Large Vision-Language Models (LVLMs). Unlike previous benchmarks that often rely on simple queries, HALLUCINOGEN introduces a diverse set of "object hallucination attacks" through complex contextual reasoning prompts. These prompts are crafted to challenge LVLMs by asking about visual objects that may or may not be present in an image, forcing the models to accurately identify, locate, or perform visual reasoning around specific objects, thereby exposing instances where they fabricate or misclassify objects. \cite{seth2024hallucinogen} 

\end{itemize}

These domain-specific benchmarks are indispensable for the safe evaluation of LLM performance in contexts where hallucinations may lead to misdiagnosis, faulty software behavior, or visual misinterpretations, thus bridging the gap between generic metrics and task-critical assessment.

\subsection{Quantitative metrics}

Metrics used to evaluate hallucination are typically classified according to the type of alignment they measure---faithfulness to input, factuality with respect to external knowledge, or semantic consistency.

\subsubsection{Faithfulness metrics}

These metrics assess whether the generated output remains consistent with the provided input or prompt:

\begin{itemize}
    \item \textbf{ROUGE, BLEU, and BERTScore:} these metrics are primarily surface-level or embedding-based similarity metrics. They evaluate the generated text by comparing it to a source or reference text, assessing shared words, phrases, or underlying semantic representations.

    \begin{itemize}
    \item \textbf{ROUGE (Recall-Oriented Understudy for Gisting Evaluation): } this metric is widely used in summarization to measure the overlap of n-grams (sequences of words) between the generated text and a reference. For example, ROUGE-1 assesses unigram overlap, ROUGE-2 measures bigram overlap, and ROUGE-L identifies the longest common subsequence. Its "recall-oriented" nature emphasizes how much of the reference's information is captured in the generated output \cite{lin2004rouge}.
    
    \item \textbf{BLEU (Bilingual Evaluation Understudy):} originally developed for machine translation, BLEU quantifies the precision of n-grams in the generated text compared to a reference, penalizing for brevity. It focuses on the extent to which the generated text's content is also present in the reference \cite{papineni2002bleu}.
    
    \item \textbf{BERTScore:} a more advanced metric, BERTScore employs contextual embeddings from large language models (such as BERT) to measure the semantic similarity between words and sentences in the generated text and the reference. This capability allows it to identify paraphrases or synonyms even when exact word matches are absent \cite{zhang2019bertscore}.
    
     \item \textbf{Limitations:}  while useful for initial assessments, these metrics are often insufficient for detecting nuanced semantic errors or fabrications. High scores on ROUGE, BLEU, or BERTScore do not guarantee factual accuracy, as a text can exhibit high lexical or semantic similarity while still containing subtle inconsistencies or outright hallucinations. They do not inherently assess the truthfulness of a statement, only its textual similarity \cite{lin2004rouge, papineni2002bleu, zhang2019bertscore}.

    \end{itemize}

    \item \textbf{FactCC:} stands for "Fact-Checking with Contextualized Commonsense," is a specialized metric developed specifically for detecting hallucinations, particularly within summarization tasks. Unlike simpler similarity metrics, FactCC employs a trained classifier. This classifier learns to identify factual inconsistencies by being trained on datasets of summaries paired with their source texts, where human annotators have meticulously labeled instances of factual inconsistencies. This methodology allows the classifier to recognize patterns that indicate unfaithfulness. Its primary advantage lies in its improved precision in hallucination detection; because it's explicitly trained to identify inconsistencies, FactCC is more effective at catching factual errors than metrics that rely solely on surface-level comparisons, aiming to determine if a generated statement truly contradicts the original source material \cite{kryscinski2019evaluating}.

    \item \textbf{SummaC:} is a metric that assesses factual consistency by leveraging the principles of Natural Language Inference (NLI). NLI is a core task in natural language processing where a model determines the logical relationship between two sentences: whether one sentence (the "hypothesis") is entailed by, contradicts, or is neutral with respect to another sentence (the "premise"). In SummaC's application, segments of the source text serve as the premise, while sentences from the AI-generated output (such as a summary) act as the hypothesis. An NLI model then evaluates if each statement in the generated output is entailed by the source text, signifying factual consistency. Conversely, statements that contradict the source or are not supported by it may indicate a hallucination or an unverified claim. SummaC's strength lies in its ability to model factual consistency through these entailment relationships, which closely align with how humans judge factual accuracy, offering a more robust assessment of faithfulness compared to methods based on simpler lexical or embedding similarity \cite{laban2022summac}.
\end{itemize}

\subsubsection{Factuality metrics}

These metrics measure the alignment of generated content with real-world facts or structured knowledge sources, moving beyond mere consistency with input prompts to verify external veracity.

\begin{itemize}
    \item \textbf{ Knowledge Intensive Language Tasks (KILT):} is a benchmark designed to evaluate the factual accuracy and knowledge-groundedness of language models. It specifically assesses the ability of generated content to align with factual information found in structured knowledge sources. The core methodology involves linking entities and claims present in the generated text to corresponding entries and facts within established knowledge bases, such as Wikipedia. This approach allows for a direct verification of whether the model's output reflects verifiable real-world knowledge rather than merely being coherent or consistent with an initial prompt. KILT's tasks often require models to generate text that can be directly verified against these external knowledge sources, making it a robust measure of a model's factual grounding \cite{petroni2020kilt}.

    \item \textbf{Retrieval-Augmented Evaluation (RAE): } is a methodology used to assess the factual grounding of generated claims, particularly within Retrieval-Augmented Generation (RAG) pipelines. RAE operates by evaluating whether the evidence retrieved by a RAG system genuinely supports the claims made in the generated output. The process typically involves two main steps: first, identifying the claims made by the language model, and second, verifying these claims against the specific knowledge snippets or documents that the retrieval component of the RAG system provided as grounding evidence. This metric offers a scalable and efficient way to judge the factual accuracy and support of generated text, as it directly checks the consistency between the model's output and its purported factual basis derived from the retrieval step. RAE is crucial for ensuring that RAG models do not hallucinate information, even when provided with relevant context \cite{lewis2020retrieval, saad2023ares, es2024ragas, salemi2024evaluating}.
\end{itemize}

\subsubsection{Human evaluation}

Despite the proliferation of automated metrics, human evaluation remains the most reliable and widely accepted method for hallucination detection. Annotators typically assess outputs based on criteria such as:

\begin{itemize}
    \item \textbf{Correctness}: this criterion assesses whether the generated content aligns with verifiable real-world facts. An output is considered correct if it can be independently validated against trusted knowledge sources. Inaccuracies, fabrications, or distortions of known information constitute a lack of correctness \cite{gekhman2023trueteacher, liu2023trustworthy}. 
    \item \textbf{Faithfulness}: measures the extent to which the model’s output remains consistent with the input prompt or source material. An output is unfaithful if it introduces information not present in the input, omits critical elements, or misrepresents the source. Faithfulness is especially important in summarization, translation, and question-answering tasks \cite{maynez2020faithfulness, chen2022towards}.
    \item \textbf{Coherence}: refers to the logical consistency and internal structure of the output. A coherent response maintains a stable topic, avoids contradictions within itself, and follows a clear and understandable flow of reasoning. Incoherent outputs may contain abrupt topic shifts, self-contradictions, or illogical argumentation \cite{parmar2024towards, beyer2021incoherence, malkin2021coherence}.
    \item \textbf{Harmfulness or bias}: this dimension captures whether the output contains content that could be ethically problematic, offensive, or unsafe. This includes outputs that propagate harmful stereotypes, generate defamatory claims, or offer misleading information in domains like medicine, law, or finance. Special attention is needed in safety-critical applications where biased or harmful content could have serious real-world consequences \cite{andriushchenko2024agentharm, zhou2023making, taubenfeld2024systematic, ling2025bias}.
\end{itemize}

However, human evaluation is costly, time-consuming, and often subject to inter-rater variability, underscoring the need for more robust and interpretable automatic metrics.

\subsection{Limitations and open challenges}

Despite advancements in automated evaluation, current benchmarks and metrics for hallucination detection in AI-generated content face several persistent limitations that impede comprehensive and comparable assessments.

\subsubsection{Lack of standardization}
A significant challenge is the absence of a universally accepted definition of hallucination. Various benchmarks and research studies adopt differing conceptualizations, leading to inconsistencies in how hallucinations are annotated and measured. This definitional variability makes it exceedingly difficult to conduct fair and meaningful comparisons of hallucination rates and detection capabilities across different models, datasets, or evaluation frameworks. The absence of a shared understanding and operationalization of "hallucination" hinders the development of generalizable solutions and a cumulative scientific discourse.

\subsubsection{Task dependence}
The effectiveness of current metrics is often highly dependent on the specific natural language processing task being evaluated. Metrics that might demonstrate reasonable performance in detecting hallucinations within summarization tasks, for instance, frequently fail to generalize or perform adequately in other domains such as question answering (QA), dialogue generation, or code generation. This limitation arises because the nature and manifestation of hallucinations can vary significantly across tasks. What constitutes a hallucination in a factual summary (e.g., inventing a detail) might differ from an unfaithful response in a dialogue system (e.g., contradicting prior turns) or an incorrect function in code generation. This task-specific performance necessitates the development of specialized metrics for each application, increasing complexity and fragmentation in the evaluation landscape.

\subsubsection{Insensitivity to subtle hallucinations}
Many existing metrics, particularly those relying on surface-level textual similarity or basic factual checks, are inherently unable to detect more nuanced and insidious forms of hallucination. These can include low-level factual shifts (slight alterations to numerical values or dates), subtle inferential errors (drawing an incorrect conclusion from otherwise correct premises), or context-dependent misalignments where a statement might be technically plausible but factually incorrect given the specific context of the input. Such subtle hallucinations are challenging to identify automatically, as they often require deep semantic understanding, complex logical reasoning, or access to vast external knowledge bases, making them particularly deceptive and hard to mitigate

\subsubsection{Limited grounding and explainability}
A critical drawback of most automatic hallucination detection scores is their lack of interpretability and diagnostic value. These metrics typically provide a numerical score indicating the presence or absence of hallucination but offer little to no insight into why a particular output is deemed hallucinated. This limited grounding means that developers receive minimal actionable feedback on the specific type of error, the source of the factual deviation, or the exact portion of the generated text responsible for the hallucination. Without this granular insight, debugging models, understanding their failure modes, and implementing targeted improvements for hallucination reduction become significantly more challenging and less efficient. The lack of explainability impedes effective model development and refinement.

\subsection{Toward unified evaluation frameworks}

Future progress depends on the development of comprehensive, taxonomy-aware, and domain-adapted evaluation frameworks that:

\begin{itemize}
    \item Incorporate \textbf{multi-level evaluation}, combining surface-level similarity with logic- and knowledge-aware assessment;
    \item Leverage \textbf{retrieval-based and symbolic tools} to enhance grounding;
    \item Standardize annotation protocols and metrics across tasks;
    \item Integrate \textbf{model uncertainty and confidence calibration} into evaluation.
\end{itemize}

Ultimately, the path to robust hallucination mitigation must be rooted in rigorous, context-sensitive measurement. Without accurate and scalable evaluation tools, efforts to reduce hallucination risk in real-world applications will remain incomplete and difficult to validate.

\vspace{0.5cm}
{\color{gray}\hrule}
\begin{center}
\section{Hallucination mitigation strategies}
\textbf{This section surveys both architectural and systemic approaches to mitigating hallucinations in LLMs, including tool augmentation, retrieval grounding, fine-tuning, symbolic guardrails, and user-facing strategies such as uncertainty displays and fallback mechanisms.}
\bigskip
\end{center}
{\color{gray}\hrule}

Given the theoretical inevitability of hallucinations in LLMs~\cite{shinn2023reflexion}, researchers and developers have proposed a range of mitigation strategies. These can be broadly categorized into two groups: architectural strategies, which modify how the model itself is trained or behaves during inference, and systemic strategies, which shape how the model is embedded, controlled, or interpreted within a broader application context. Both are necessary for creating robust, trustworthy systems that minimize the frequency and harm of hallucinations.

\subsection{Architectural mitigation strategies}

Architectural strategies operate at the model level and seek to reduce hallucination risk by directly improving the model’s grounding, reasoning, or factual alignment capabilities. These interventions are typically implemented through changes in training data, model design, or auxiliary components used at inference time.

\subsubsection{Toolformer-style augmentation}

Recent advances in tool-augmented LLMs propose allowing the model to call external APIs, calculators, code interpreters, or structured knowledge tools during inference. Toolformer~\cite{schick2023toolformer}, for example, fine-tunes an LLM to decide when and how to use external tools to answer questions more reliably. Instead of relying purely on parametric memory, the model delegates subtasks—such as date calculations, currency conversions, or fact retrieval—to external systems better equipped to handle them.

This approach offloads fact-intensive or computation-heavy tasks to specialized modules, significantly reducing hallucination risk in those areas. The model learns to invoke tools autonomously during generation, producing more grounded and verifiable responses while maintaining fluency.

\subsubsection{Factual grounding through retrieval mechanisms}

\textit{(RAG)} is one of the most widely adopted hallucination mitigation frameworks. In RAG systems~\cite{lewis2020retrieval, izacard2020leveraging}, the model is paired with a retrieval component that fetches relevant documents or knowledge snippets from a curated corpus (e.g., Wikipedia, academic papers, or enterprise databases) in response to a user query. These retrieved documents are then passed as additional input context to the LLM, grounding the generated response in verifiable sources.

RAG reduces the likelihood of hallucination by:
\begin{itemize}
    \item Constraining generation to content retrieved from external knowledge bases.
    \item Allowing the model to quote, summarize, or paraphrase from known references.
    \item Providing users with transparency and traceability via document citations.
\end{itemize}

Prominent RAG implementations include Google's Bard, Meta's BlenderBot 3, and enterprise systems like Microsoft’s Copilot and Amazon Bedrock. Despite its advantages, RAG is not foolproof: the model may still hallucinate if it fails to properly interpret or align with the retrieved material~\cite{shuster2021retrieval}.

\subsubsection{Fine-tuning with synthetic or adversarially filtered data}

Another mitigation strategy involves fine-tuning LLMs on curated or synthetic datasets designed to reduce hallucination tendencies. Two prominent approaches include:

\begin{itemize}
    \item \textbf{Synthetic factuality tuning}: Models are trained or fine-tuned on large corpora of verified, well-grounded question-answer pairs. These may be created through human annotation or automatically generated and filtered using factual consistency metrics~\cite{honovich2022true}.
    
    \item \textbf{Adversarial filtering}: Using hallucination detection models or adversarial prompts to identify and filter out outputs that exhibit hallucination. These filtered outputs can be used to refine the LLM or train classifier modules that flag likely hallucinated content~\cite{le2020adversarial}.
\end{itemize}

Although effective in reducing hallucinations on benchmark tasks, these methods face limitations in scalability, domain generalizability, and susceptibility to dataset bias.

\subsection{Systemic mitigation strategies}

Systemic strategies are applied at the deployment or user interface level and focus on shaping how LLM outputs are interpreted, controlled, or constrained in real-world contexts. These strategies often complement architectural solutions by providing guardrails and transparency mechanisms to reduce the risk and impact of undetected hallucinations.

\subsubsection{Guardrails and symbolic integration}

Guardrails are rule-based or symbolic control mechanisms that constrain the behavior of LLMs during inference. These include:

\begin{itemize}
    \item \textbf{Logic validators}: these components evaluate whether an LLM's output is internally consistent or conforms to formal rules in a given domain—such as mathematics, programming, or natural language logic. For example, in arithmetic tasks, a logic validator can compare the model's answer against a rule-based calculator. In legal or contractual reasoning, outputs can be assessed for compliance with regulatory clauses or logical structures. By acting as a correctness gatekeeper, logic validators help identify outputs that may be fluent but logically invalid \cite{albrightimproving}.

    \item \textbf{Factual filters}: factual filtering involves post-processing model outputs to detect contradictions or inconsistencies with a trusted external source, such as a structured database or knowledge graph (e.g., Wikidata, UMLS, or proprietary enterprise data). These systems can match generated claims against canonical facts and either flag inaccuracies or attempt automatic correction. For instance, if a model claims that ``Paris is the capital of Germany,'' a factual filter could detect the mismatch and suggest a correction based on structured geopolitical data~\cite{dong2024building}. Such filters are particularly valuable in domains where factual consistency is non-negotiable, like medicine, finance, and policy generation.

    \item \textbf{Rule-based fallbacks}: in scenarios where uncertainty is high or outputs are flagged as potentially hallucinated—either by a validator, confidence threshold, or user feedback—the system can execute predefined fallback policies. These include refusing to answer (e.g., ``I'm not confident enough to provide a reliable response''), rerouting the request to a human-in-the-loop, or prompting the user for clarification. Rule-based fallbacks act as safety valves, especially in high-stakes contexts, by enforcing cautious behavior when confidence or factuality cannot be guaranteed. They are also used in frameworks such as Reinforcement Learning with Human Feedback (RLHF), where such flags inform future model training \cite{li2025drift}.
\end{itemize}

Symbolic integration—where models are combined with deterministic reasoning systems—represents a promising frontier for hallucination mitigation. Neuro-symbolic systems, for example, blend statistical generation with formal logic, enabling models to verify or revise outputs before presentation~\cite{mao2019neuro}.

\subsection{Toward hybrid and context-aware mitigation systems}

As no single technique fully eliminates hallucinations across all tasks and domains, the most promising direction lies in the development of hybrid mitigation systems—architectures that combine multiple complementary strategies to reduce both the frequency and the impact of hallucinated outputs.

An effective hybrid system integrates strengths from various approaches:
\begin{itemize}
    \item \textbf{Tool use} enables precise, verifiable outputs by delegating computation-heavy or fact-specific tasks (e.g., date calculation, code execution, currency conversion) to external APIs or structured functions.
    
    \item \textbf{Retrieval grounding} supplements the model’s internal representations with up-to-date and verifiable information drawn from external sources, reducing reliance on the model’s imperfect parametric memory.
    
    \item \textbf{Fine-tuning} shapes the model's inductive biases, helping it learn patterns of truthfulness and factual consistency based on curated datasets or adversarially filtered examples.
    
    \item \textbf{Guardrails}—such as rule-based filters, logic validators, or knowledge-based correction systems—enforce hard constraints and provide a safety net to catch hallucinations that might bypass other safeguards.
\end{itemize}

In addition to being hybrid, future mitigation systems should be context-aware, meaning they adapt their strategies dynamically based on the specifics of the application. For example:
\begin{itemize}
    \item In \textit{high-stakes domains} like medicine, law, or finance, the system may be configured to prioritize factual accuracy over fluency, enforce mandatory citation or retrieval grounding, and invoke fallback procedures when confidence is low.
    \item In \textit{creative or exploratory domains} such as brainstorming or storytelling, the system might allow more open-ended generation with relaxed factual constraints, while still flagging potentially unverifiable claims.
    \item In \textit{user-facing applications}, personalization mechanisms could adjust how hallucination warnings, uncertainty indicators, or references are displayed based on user expertise or preferences.
\end{itemize}

In summary, while hallucination is an inherent risk in current-generation LLMs, its harm can be significantly reduced through a layered approach that combines architectural improvements with systemic controls tailored to the use case. Hybrid, context-sensitive systems represent a practical and responsible path toward building language models that are not only powerful, but also trustworthy, accountable, and safe for deployment in real-world environments.

\vspace{0.5cm}

{\color{gray}\hrule}
\begin{center}
\section{Monitoring LLM releases and performance: web-based resources}
\textbf{This section introduces leading web-based resources that researchers can use to monitor LLM releases and evaluate model performance over time, including hallucination trends, intelligence benchmarks, user preferences, and system cost.}
\bigskip
\end{center}
{\color{gray}\hrule}

As LLMs evolve rapidly, staying informed about new releases, performance metrics, and emerging trends is crucial for researchers, developers, and decision-makers. Rather than relying on static performance comparisons, the following platforms offer continuously updated, transparent, and publicly accessible dashboards that track and evaluate LLM capabilities across a range of tasks, including reasoning, factuality, speed, cost, and hallucination rates.

\subsection{Artificial Analysis }

Artificial Analysis \footnote{\url{https://artificialanalysis.ai}} is a comprehensive, independent benchmarking platform that compares LLMs and other AI models across several key dimensions. This resource is particularly useful for tracking models that prioritize reasoning and factual grounding—important proxies for hallucination control—even if hallucination itself is not directly measured. Its core features include.

\subsubsection{Intelligence index}
A composite score based on multiple benchmarks (e.g., MMLU-Pro, GPQA, HLE, LiveCodeBench) reflecting reasoning, problem-solving, and factual capabilities.

\begin{figure}[H]
\centering
\includegraphics[width=0.9\textwidth]{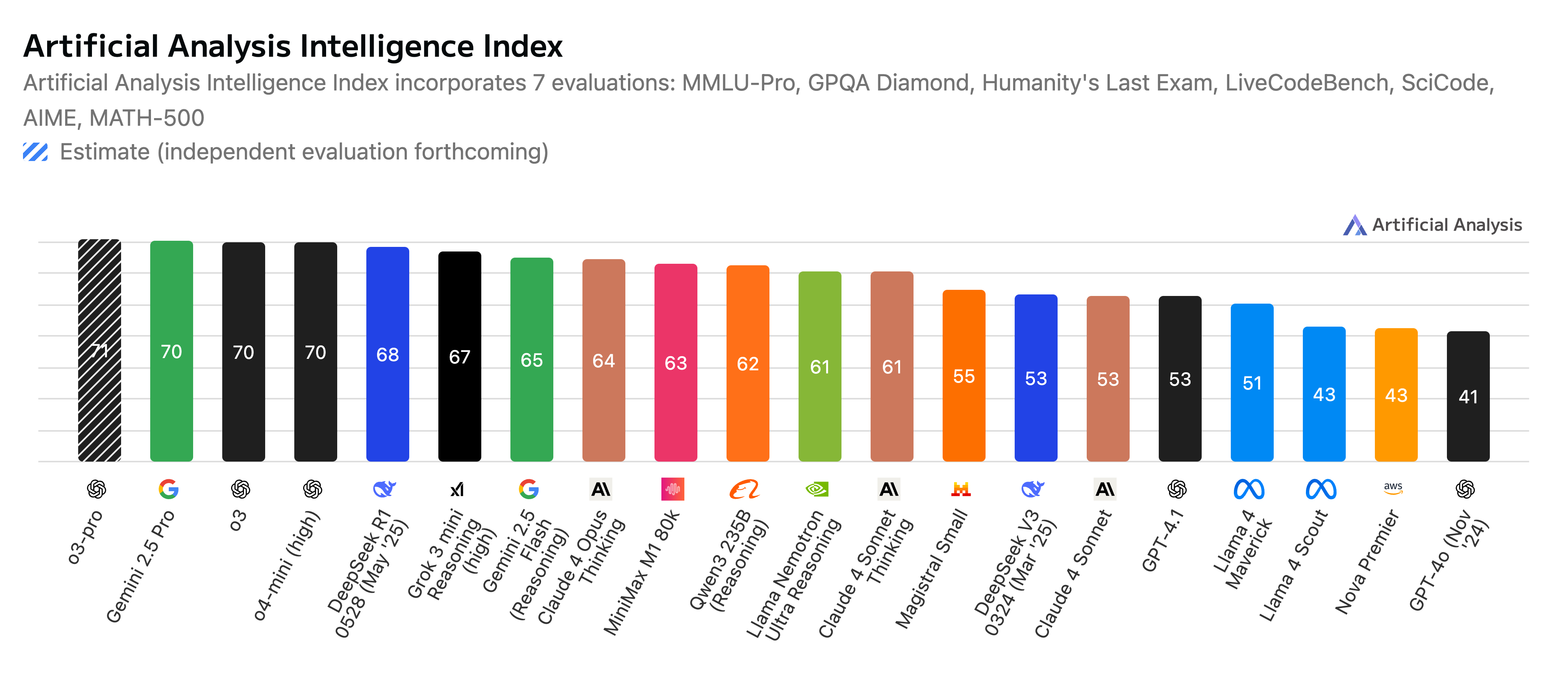}
\caption{Sample visualization of the AI Index, retrieved on 28 June 2025}
\end{figure}

\subsubsection{Cost and latency}

Side-by-side comparisons of price per 1K tokens, response latency, and context window limits for models across different providers.

\begin{figure}[H]
\centering
\includegraphics[width=0.9\textwidth]{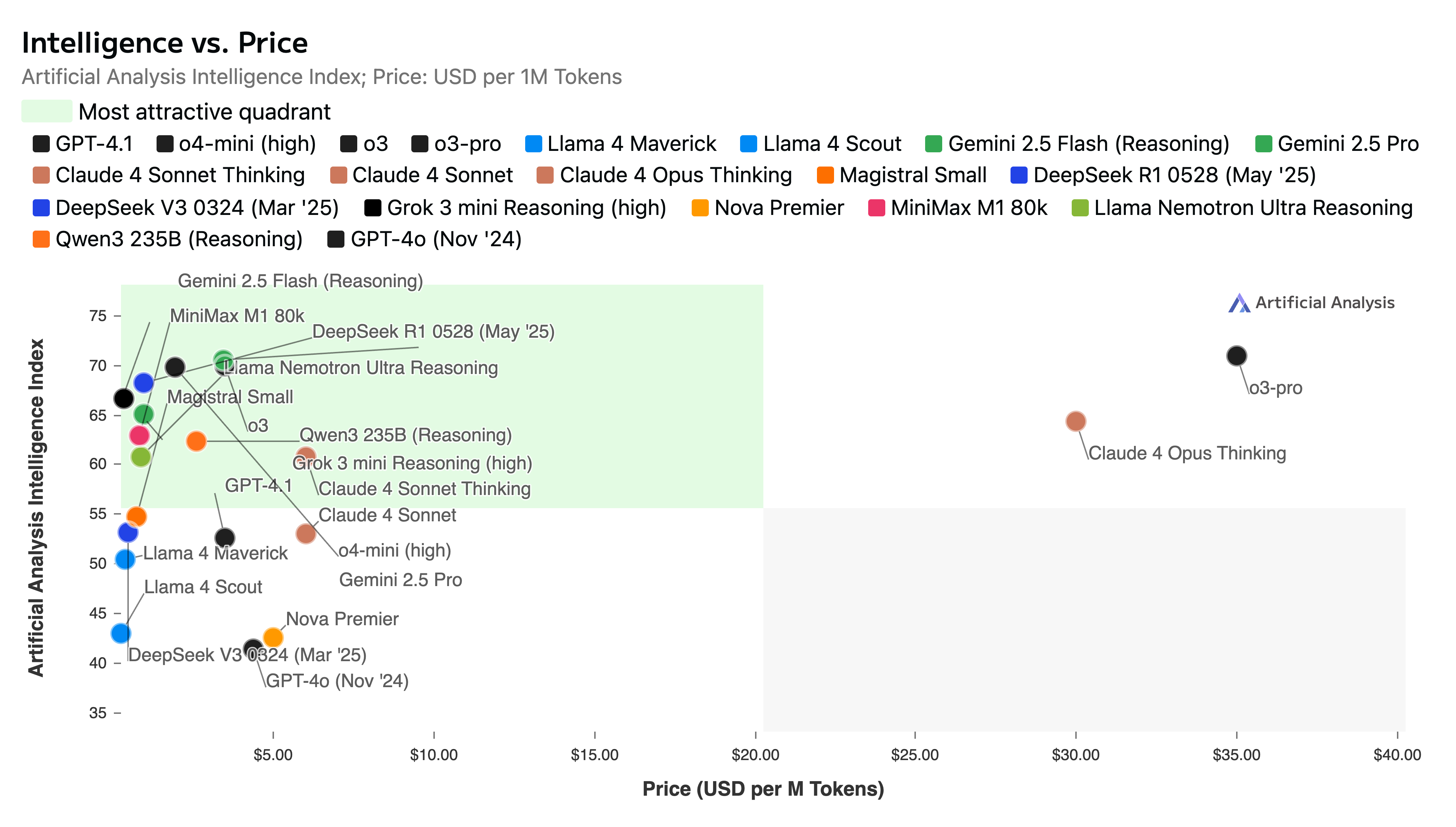}
\caption{Sample visualization of intelligence versus price, retrieved on 28 June 2025}
\end{figure}

\begin{figure}[H]
\centering
\includegraphics[width=0.9\textwidth]{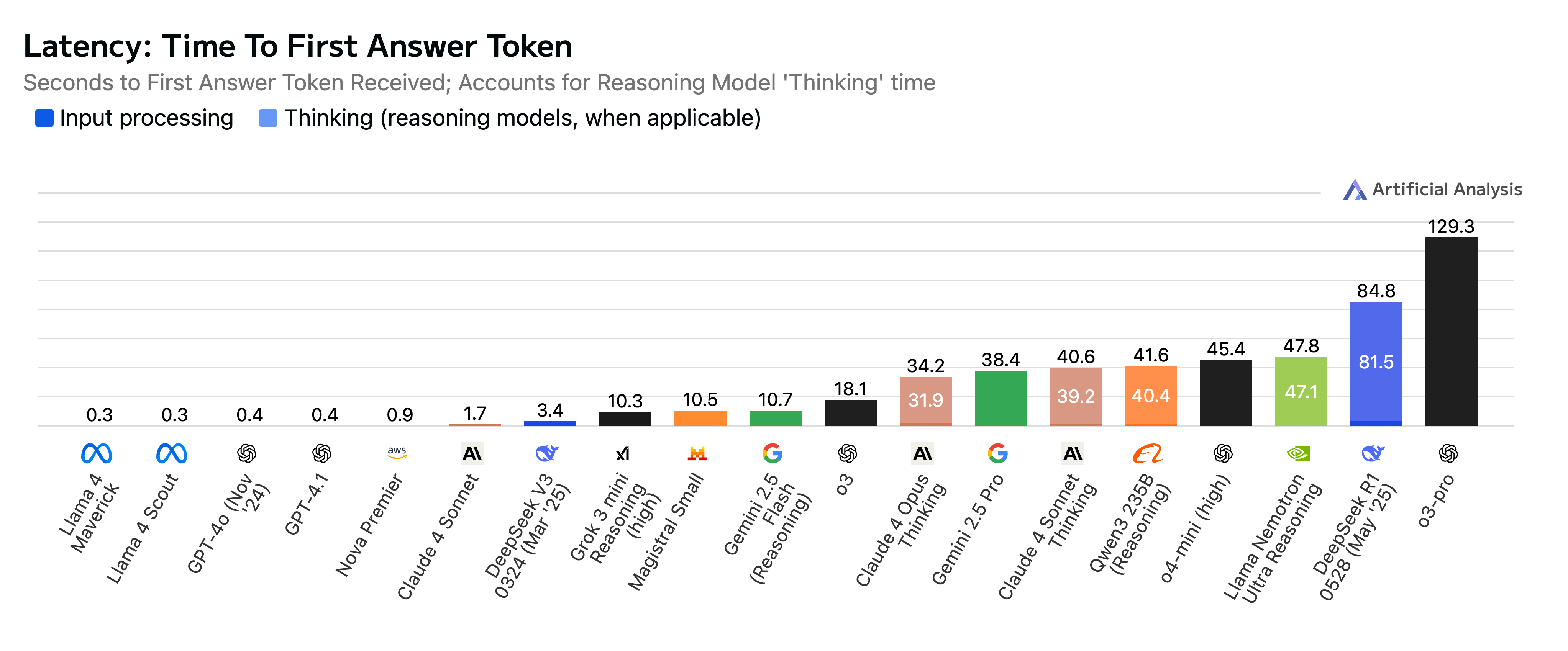}
\caption{Sample visualization of latency, retrieved on 28 June 2025}
\end{figure}

\subsubsection{Model pages}
In-depth performance breakdowns for each model version (e.g., GPT-4o, Claude 3.5, LLaMA 3.1), including context sensitivity, update history, and comparative strengths.

\subsubsection{Multimodal and API benchmarks}
Performance data for text-to-image, code, audio, and video generation models.

\begin{figure}[H]
\centering
\includegraphics[width=0.9\textwidth]{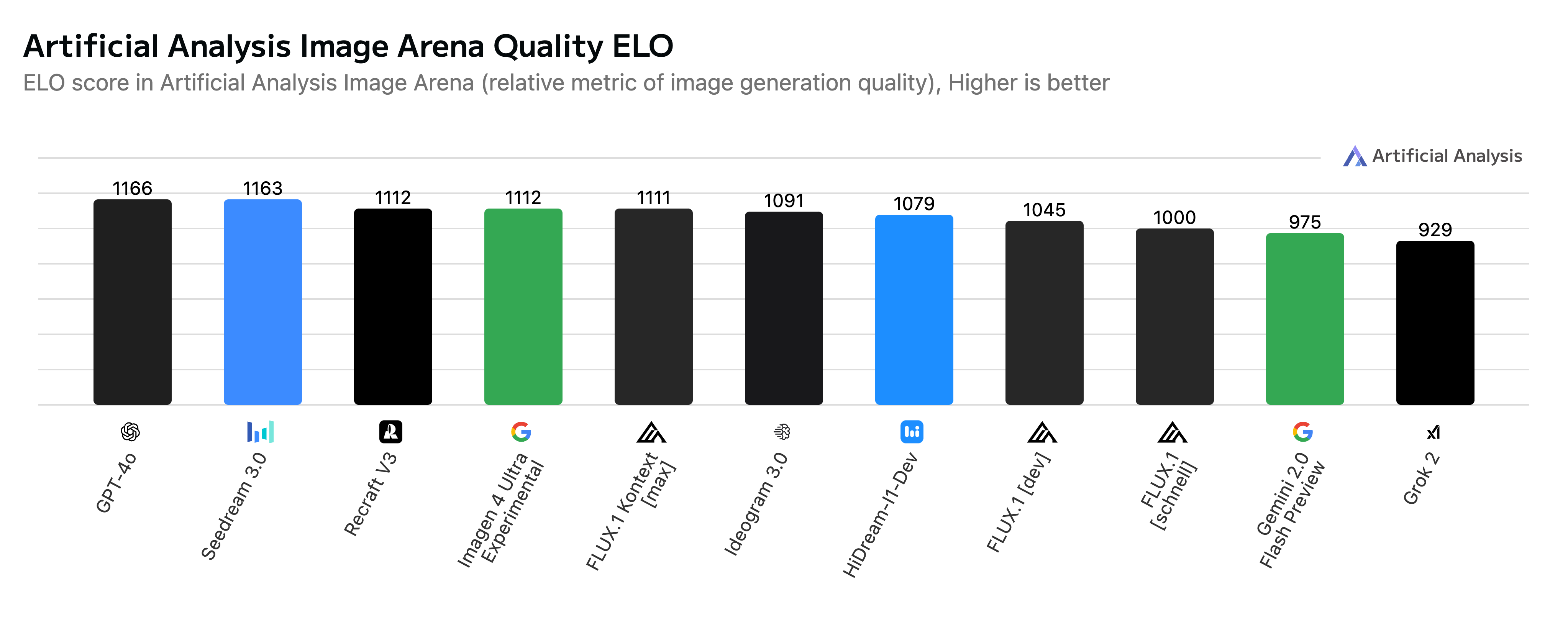}
\caption{Sample visualization of text to image, retrieved on 28 June 2025}
\end{figure}

\begin{figure}[H]
\centering
\includegraphics[width=0.9\textwidth]{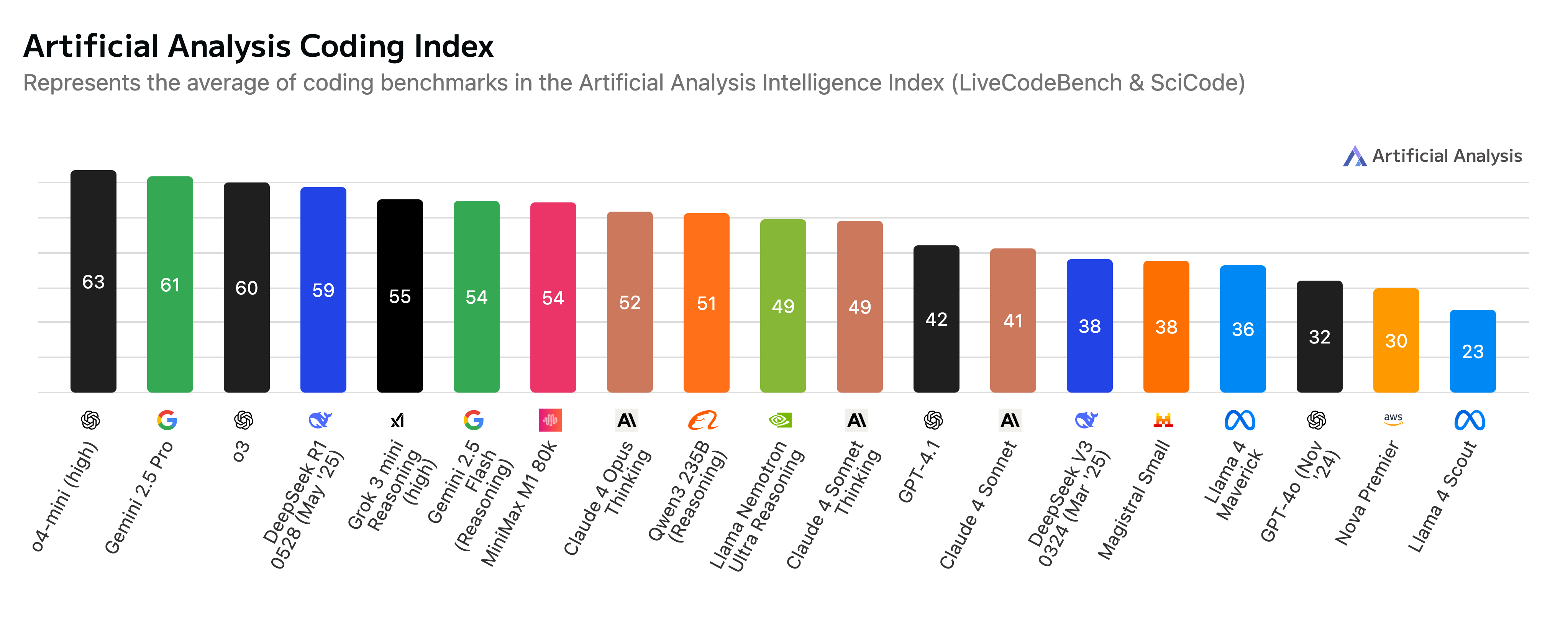}
\caption{Sample visualization of code generation, retrieved on 28 June 2025}
\end{figure}

\begin{figure}[H]
\centering
\includegraphics[width=0.9\textwidth]{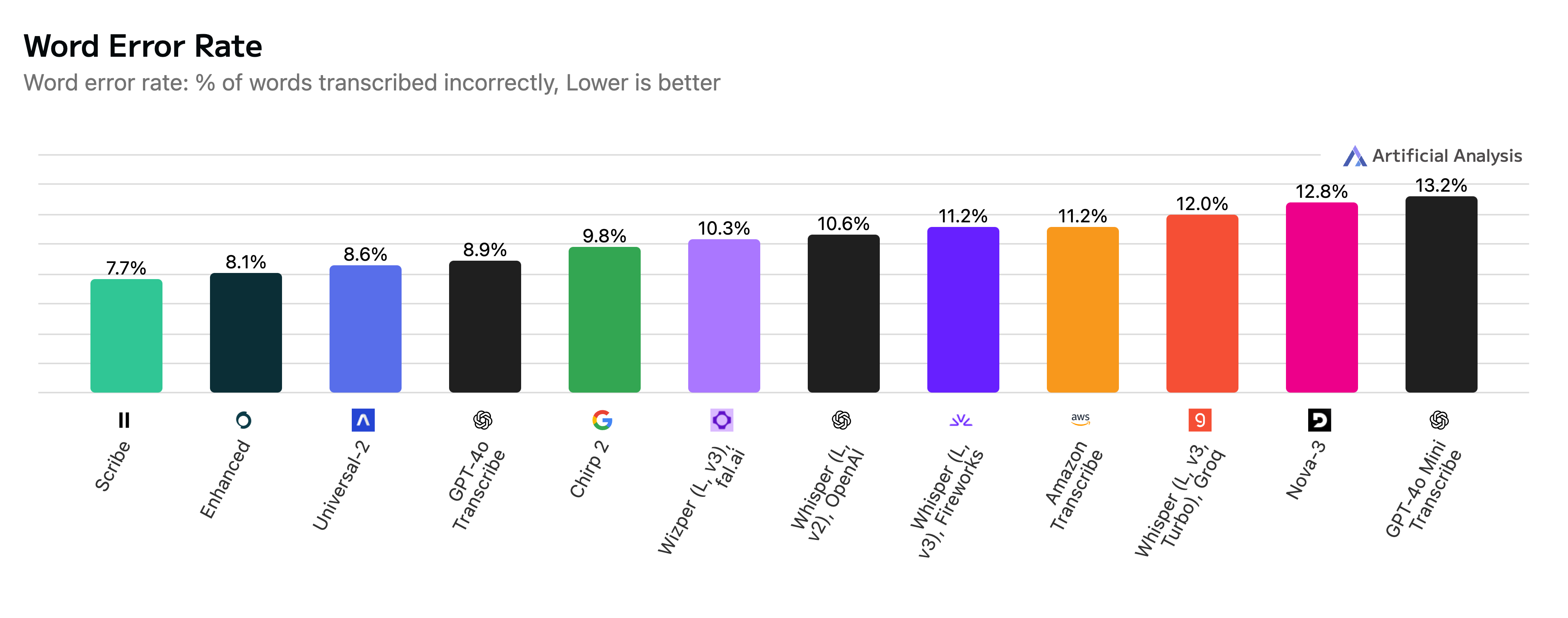}
\caption{Sample visualization of audio generation (text to speech, word error rate), retrieved on 28 June 2025}
\end{figure}

\subsubsection{Quarterly state-of-AI reports}
Strategic summaries and macro-level trends such as architectural shifts, emergent capabilities, and safety trade-offs.

\begin{figure}[H]
\centering
\includegraphics[width=0.9\textwidth]{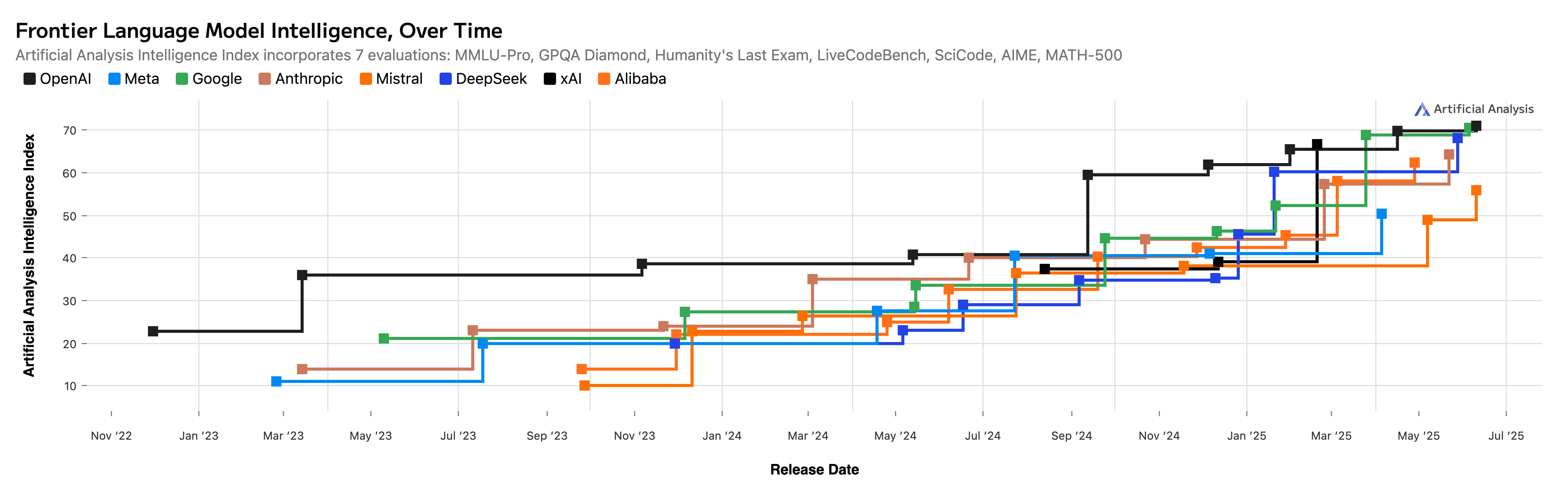}
\caption{Sample visualization of intelligence over time, retrieved on 28 June 2025}
\end{figure}

\subsection{Vectara Hallucination Leaderboard}

Maintained by Vectara, this GitHub-based leaderboard\footnote{\url{https://github.com/vectara/hallucination-leaderboard}} is one of the few public efforts that explicitly tracks \textit{hallucination rates} in LLM outputs. Its key features include:

\begin{itemize}
    \item \textbf{Task-specific evaluation:} focuses on hallucination in summarization tasks using human-labeled ground truth references.
    \item \textbf{Quantitative metrics:} reports both hallucination rate (percentage of hallucinated responses) and accuracy rate (non-hallucinated responses), usually based on hundreds of test samples per model.
    \item \textbf{Comparative analysis:} includes OpenAI models (GPT-3.5, GPT-4), Anthropic’s Claude, Meta’s LLaMA, and other open-source models.
    \item \textbf{Consistent benchmarking methodology:} uses the same prompt structure and evaluation rubric across all models for fair comparison.
\end{itemize}

This leaderboard is an essential tool for researchers focusing on the hallucination problem, particularly in summarization-heavy applications such as legal, academic, or news content generation.

\begin{figure}[H]
\centering
\includegraphics[width=0.9\textwidth]{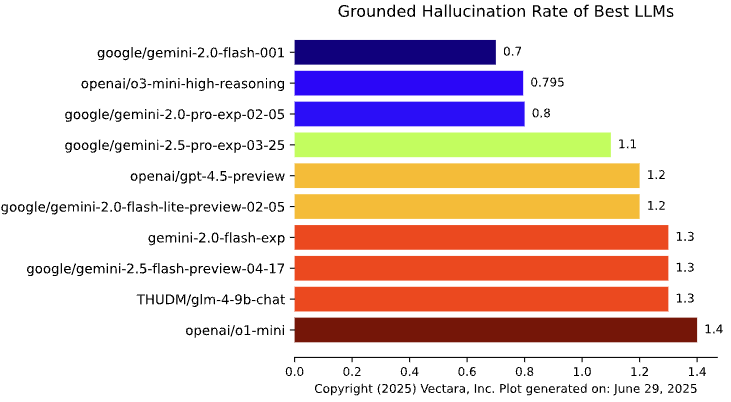}
\caption{Sample visualization of grounded hallucinations rate using Hughes hallucination evaluation model, retrieved on 29 June 2025}
\end{figure}

\subsection{Epoch AI Benchmarking Dashboard}

The Epoch AI Benchmarking Dashboard\footnote{\url{https://epoch.ai/data/ai-benchmarking-dashboard}}, a project by the nonprofit research organization Epoch, serves as a valuable resource for understanding the long-term trends in AI capabilities. While it doesn't directly measure hallucination, the dashboard's focus on benchmarks such as factual QA and reasoning indirectly reflects a model's propensity for generating inaccurate or fabricated information, making it useful for meta-analyses on the topic. By understanding the factors that contribute to higher accuracy in these benchmarks, we can infer strategies for mitigating hallucinations.

The dashboard offers several key features that contribute to this understanding:
\begin{itemize}
\item \textbf{Cross-time comparisons:} it meticulously tracks the improvement of leading models over time across various tasks, including MMLU, ARC, and BIG-bench. Improvements in these areas, particularly factual recall and logical consistency, directly correlate with a reduced likelihood of hallucination.
 \item \textbf{Benchmark aggregation:} it combines results from reasoning, language understanding, and coding tasks, enabling a comprehensive analysis of broad capability trends. Enhanced reasoning and language understanding are critical for models to accurately interpret prompts and generate contextually relevant, non-hallucinatory responses.
\item \textbf{Historical model context:} the dashboard incorporates models dating back to 2018, providing a rich historical perspective on performance scaling and the evolution of AI systems. This allows for observing how changes in model architecture and training have impacted accuracy, offering insights into what contributes to more reliable outputs.
\item \textbf{Data transparency:} it prioritizes transparency by including information on dataset origins, training scale, and publicly available model sizes. This transparency is crucial for researchers to understand the underlying factors that contribute to model performance and, by extension, to identify potential sources of hallucination.
\end{itemize}

Beyond these general features, the provided graphs and accompanying text explanations from Epoch AI highlight several insightful trends pertinent to AI performance and, by extension, offer direct implications for combating hallucinations:

\subsubsection{Accuracy versus training compute}
As illustrated in the "Accuracy versus training compute" graph, a clear correlation exists between the estimated training compute (in FLOPs) and the GPQA Diamond and MATH Level 5 accuracies. For GPQA Diamond, models with less than $10^{24}$ FLOPs often struggle to perform above random chance, sometimes even performing worse due to difficulties in understanding question formatting. This indicates a higher likelihood of generating nonsensical or hallucinated answers. However, beyond this threshold, performance shows a notable increase of approximately 12 percentage points for every 10x increase in compute. Similarly, on MATH Level 5, models with higher compute estimates generally achieve higher scores, with performance increasing around 17 percentage points for every 10x increase in pretraining compute, though this trend appears noisier.

The direct implication for fighting hallucinations is clear: increased training compute leads to more accurate models, which are inherently less prone to hallucinate. Greater computational resources allow models to learn more intricate patterns, better understand factual relationships, and perform more robust reasoning, all of which directly combat the generation of fabricated information. The graph also underscores the impact of algorithmic progress: more recent models like DeepSeek-R1, Phi-4, or Mistral Small 3 surpass older models trained with comparable compute. This suggests that algorithmic advancements, alongside increased compute, are vital for developing models that are more reliably accurate and thus less prone to hallucinate.

\begin{figure}[H]
\centering
\includegraphics[width=0.9\textwidth]{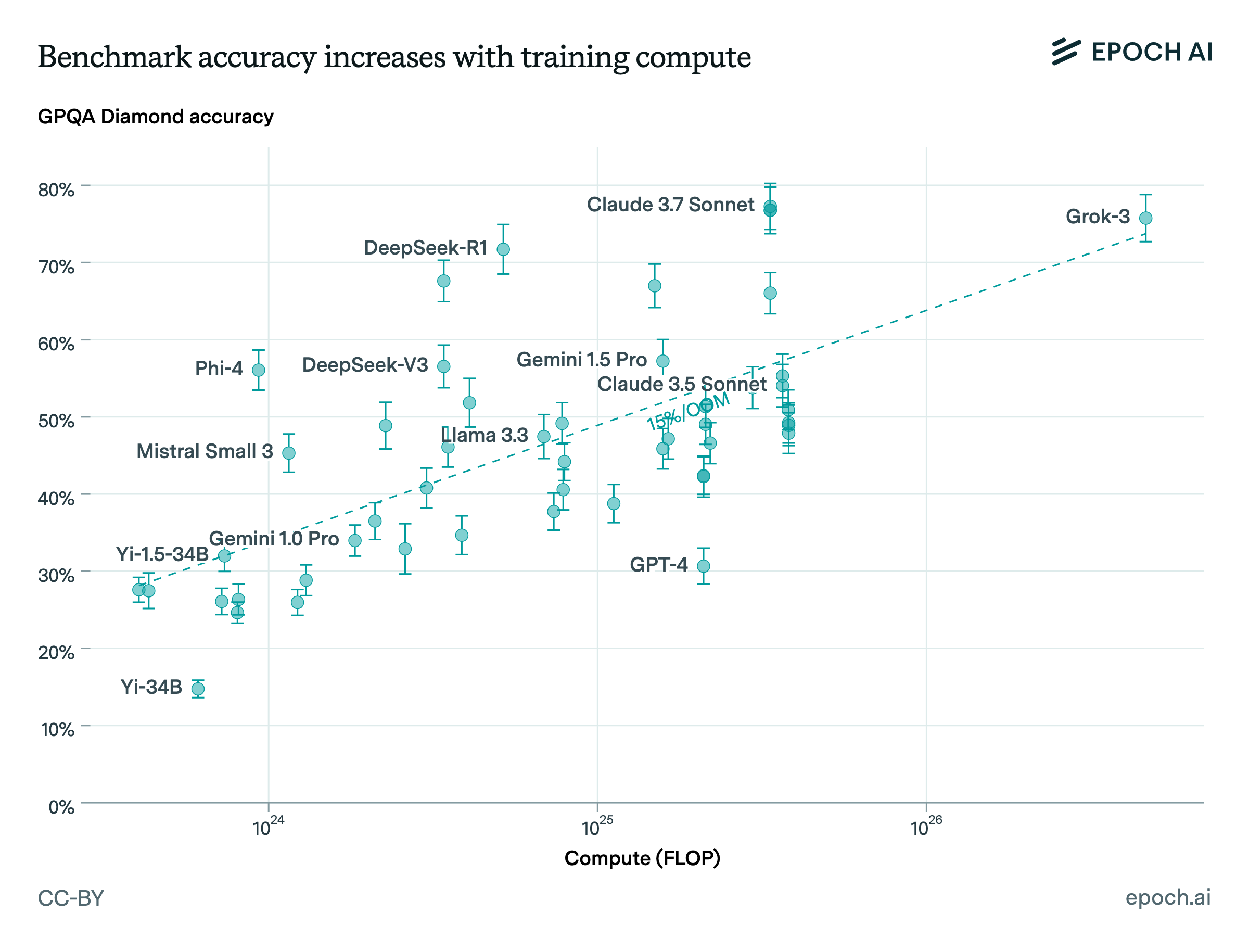}
\caption{Sample visualization of accuracy versus training compute, retrieved on 29 June 2025}
\end{figure}

\subsubsection{Open source versus propietary}
The "Models with downloadable weights vs proprietary" graph demonstrates a discernible performance gap between models with downloadable weights (often open-source) and their top-performing, often proprietary, counterparts. On the GPQA Diamond benchmark, models with downloadable weights tend to lag behind. For instance, in January 2025, OpenAI’s o1 outperformed the best downloadable model at the time, Phi-4, by a significant 20 percentage points. A similar disparity was observed on MATH Level 5, where Phi-4 trailed o1 by 29 percentage points. Epoch's analysis further suggests that the best-performing open LLMs lagged the best closed LLMs by a considerable margin, ranging from 6 months on GPQA Diamond to 20 months on MMLU.

This performance gap has significant implications for addressing hallucinations. The limited access to the weights of top-performing models hinders open research into the root causes and mitigation strategies for hallucinations. Researchers cannot directly probe or modify these models to understand why they generate more accurate (and thus less hallucinatory) outputs. However, the release of DeepSeek-R1 in January 2025 marked a notable shift, significantly narrowing this performance gap. On MATH Level 5, DeepSeek-R1 only lagged behind the then-best-performing model, o3-mini, by 2 percentage points. This closing gap is crucial as it suggests that open-source models are catching up in terms of accuracy, which could accelerate community-driven efforts to understand and reduce hallucinations in publicly available models.

\begin{figure}[H]
\centering
\includegraphics[width=0.9\textwidth]{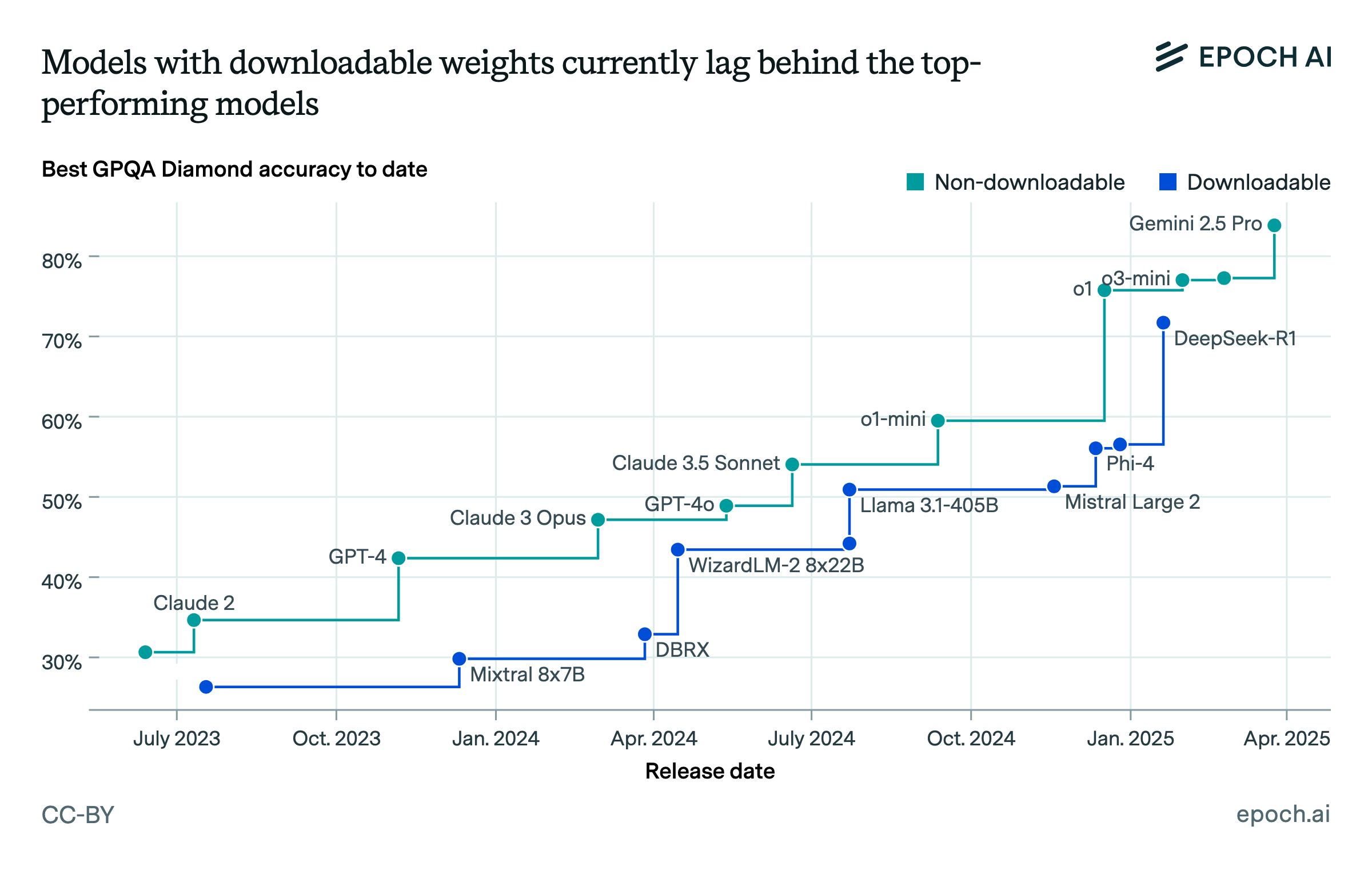}
\caption{Sample visualization of models with downloadable weights vs proprietary, retrieved on 29 June 2025}
\end{figure}

\subsubsection{Geographic disparities and performance}
The "US models vs non-US" graph highlights a consistent trend where leading US-developed models exhibit higher accuracies than non-US models on both GPQA Diamond and MATH Level 5 benchmarks. For example, OpenAI’s o1 leads on GPQA Diamond, while o3-mini holds the top spot on MATH Level 5, both being US models.

This geographical disparity in performance indirectly impacts the global effort to combat hallucinations. Higher-performing models, regardless of their origin, generally exhibit fewer hallucinations due to their superior understanding and reasoning capabilities. The current dominance of US models suggests that much of the cutting-edge research driving accuracy improvements (and thus hallucination reduction) might be concentrated in US-based organizations. However, with the release of DeepSeek-R1 in January 2025, the performance gap between US and non-US models has substantially reduced. DeepSeek-R1, a non-US model, now trails o3-mini by only 2 percentage points on MATH Level 5 and scores just 4 percentage points lower than o1 on GPQA Diamond. This narrowing gap indicates a more distributed advancement in AI capabilities, which is positive for fostering diverse approaches and collaborations in the fight against hallucinations globally.

\begin{figure}[H]
\centering
\includegraphics[width=0.9\textwidth]{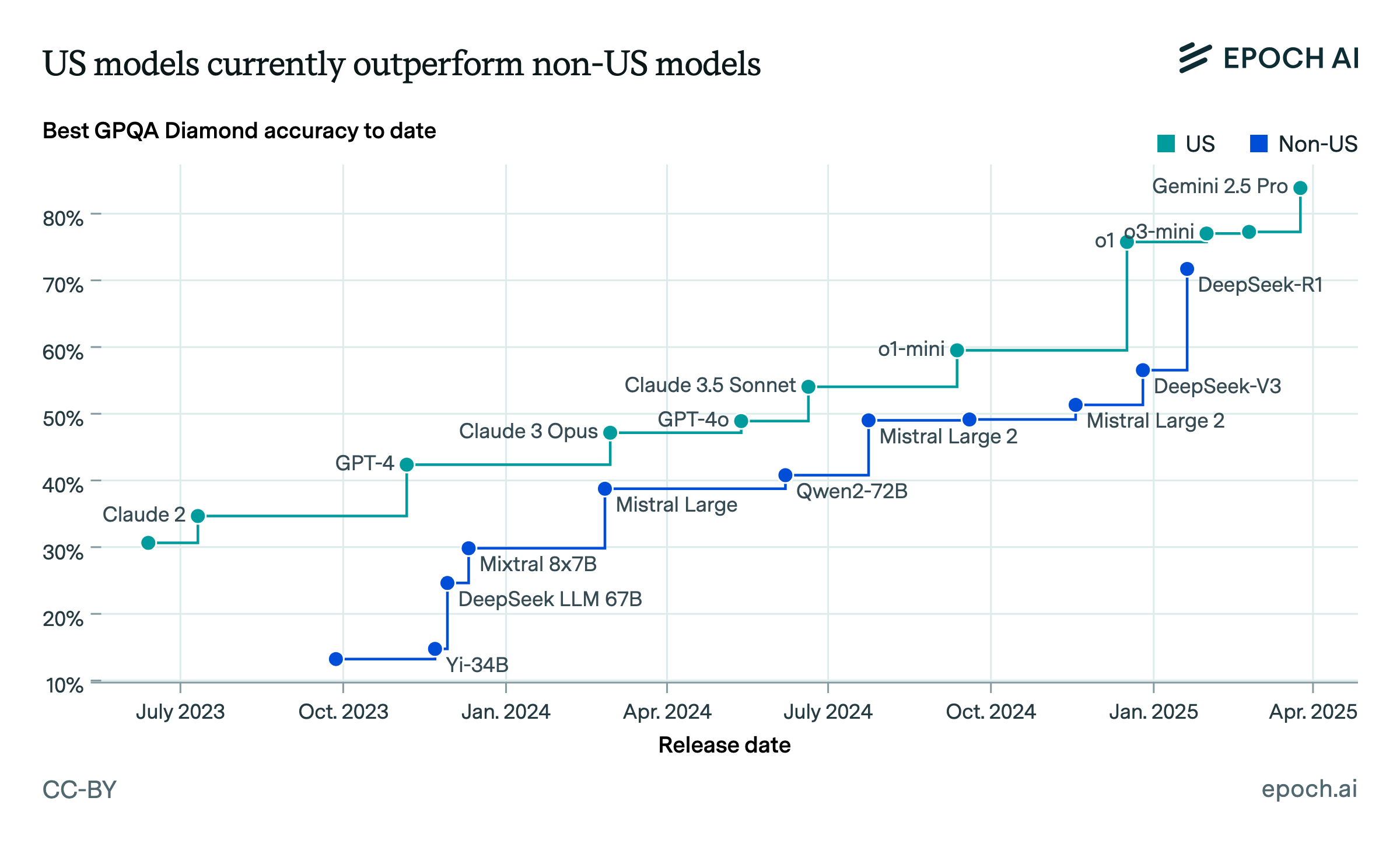}
\caption{Sample visualization of US models vs non-US, retrieved on 29 June 2025}
\end{figure}

\subsubsection{Performance on expert-level mathematics problems}
The "models performance on expert-level mathematics problems" graph, focusing on FrontierMath accuracy, provides insights into the capabilities of various AI models in tackling complex mathematical challenges. While the previous graphs mainly used GPQA Diamond, this one provides a different perspective on performance. The data points show the FrontierMath accuracy for different models from various organizations (OpenAI, Anthropic, xAI, Google, Mistral AI, Alibaba, Meta AI, DeepSeek) over time, with error bars indicating the range of results. Notably, models like o4-mini (medium), o3-mini (high), and o1 (high) from OpenAI demonstrate some of the highest accuracies, particularly in the later part of the timeline (late 2024 to early 2025).

This specific benchmark, while not directly tied to hallucination, is crucial for assessing a model's logical reasoning and problem-solving abilities. A high degree of accuracy on expert-level mathematics problems is a strong indicator of a model's foundational understanding and ability to produce precise, non-contradictory outputs. Models that struggle with such tasks are more likely to generate illogical or fabricated results in less constrained domains. Therefore, improvements in FrontierMath accuracy can be seen as a proxy for increased robustness against hallucinations, as it signifies a deeper and more reliable cognitive capacity within the AI model.

\begin{figure}[H]
\centering
\includegraphics[width=0.9\textwidth]{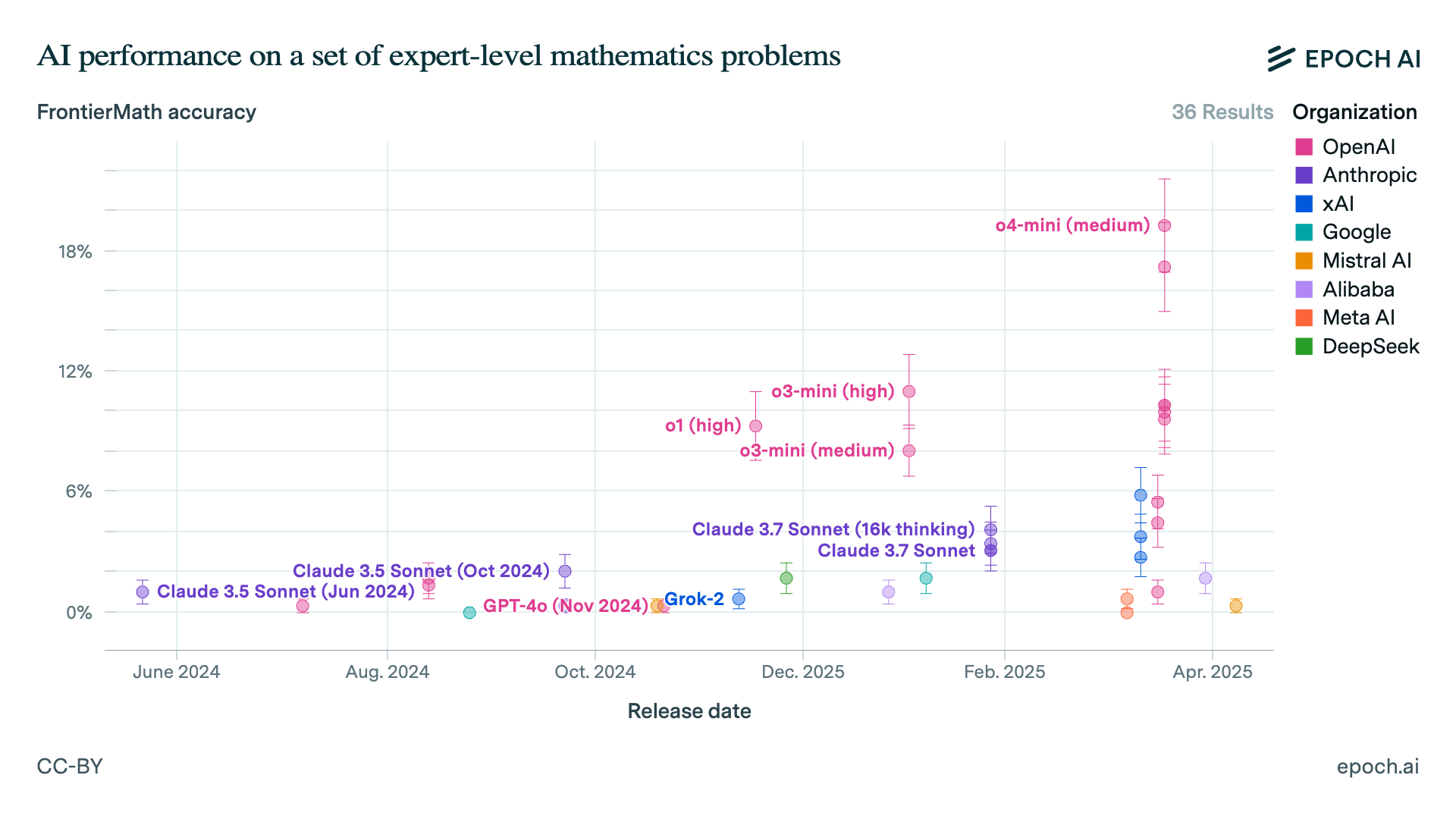}
\caption{Sample visualization of models performance on expert-level mathematics problems, retrieved on 29 June 2025}
\end{figure}

\subsection{LM Arena}

Created by researchers from UC Berkeley’s SkyLab, LM Arena (formerly LMSYS Chatbot Arena)\footnote{ \url{https://lmarena.ai/leaderboard}} is an open platform where anyone can easily access, explore, and interact with the world’s leading AI models. The platform's foundational values, as expressed by its founders when the first leaderboard was launched in July 2023, are deeply rooted in research: to create a rigorous, reproducible, community-led framework for real-world model evaluation. This commitment to open, community-driven assessment makes LM Arena a uniquely valuable resource for understanding AI model behavior, particularly concerning the critical issue of hallucination. While not explicitly designed as a hallucination detector, its methodology provides crucial qualitative insights into how users perceive model accuracy, helpfulness, and trustworthiness, all of which are intrinsically linked to the presence or absence of factual inaccuracies or fabricated information.

The core of LM Arena's operation, designed like a tournament, involves models being compared anonymously side-by-side, with users voting for the better response. This structure of anonymous battles, dynamic prompts (approximately 70\% of prompts each month are fresh), and a rotating user base, was specifically designed to reduce bias and reflect diverse, real-world use cases. This dynamic testing environment makes it impossible for models to predict or "memorize" specific evaluation scenarios, ensuring that their performance, including their tendency to hallucinate, is genuinely assessed based on real-time, novel interactions. Its key features are particularly relevant to the study of hallucinations:

\subsubsection{Battle-style comparisons and dynamic prompts} 
Models are compared head-to-head in blind A/B testing formats by real users, who vote on preferred responses. The anonymity of models and the constant introduction of fresh prompts mean that models cannot optimize for specific test cases. This structure means users are implicitly evaluating which model provides more accurate and reliable information, directly penalizing models that hallucinate. A user is far more likely to prefer a response that is factual, coherent, and contextually appropriate over one that contains made-up details or nonsensical statements. This "real-world usage" approach contrasts sharply with static benchmarks, providing a robust signal for hallucination.

\subsubsection{Diversity of models and transparent testing} 
The platform includes a wide array of both proprietary (e.g., OpenAI, Anthropic, Google, Meta, Alibaba) and open-source (e.g., Mistral, LLaMA, Zephyr) models, with over 40\% of battles involving an open model. LM Arena works directly with model providers for testing, comparing, and improving models both before (via pseudonyms/codenames) and after official release. This provides a shared infrastructure for reproducible and transparent evaluation. This diverse representation and testing methodology allows for comparisons across different development philosophies and architectures, helping to identify whether certain approaches or model types are inherently more prone to hallucination in various real-world interaction contexts. Only publicly released models with longer-term support get ranked on the leaderboard, ensuring that the community can verify results through their own testing.

\subsubsection{High-quality qualitative judgments driven by intrinsic motivation} 
Crucially, voters implicitly assess fluency, factuality, helpfulness, and hallucination tendencies—providing a rich complementary signal to traditional benchmarks. Since there's no payment or external incentive, votes come from intrinsic motivation, fostering a community of diverse subject-matter experts who provide authentic, thoughtful evaluations on their own prompts. When a user deems a response unhelpful, untrustworthy, or simply incorrect, it often stems from a factual inaccuracy or a fabricated piece of information—a direct manifestation of hallucination. These high-quality qualitative judgments, gathered at scale, offer a powerful "human-in-the-loop" feedback mechanism for identifying models that consistently produce non-hallucinatory content. They capture the nuanced user experience and reactions to model outputs that purely quantitative metrics might miss.

\subsubsection{Community-shaped leaderboard and transparency} 
Over time, these votes in battle mode add up to a public leaderboard that reflects collective, real-world judgment. This democratically shaped leaderboard makes AI progress more transparent, accessible, and grounded in actual usage. Models consistently ranking higher are, by extension, those perceived as more reliable and less prone to frustrating users with incorrect or fabricated information. This dynamic ranking can serve as an early warning system for models that start exhibiting higher hallucination rates, as user dissatisfaction would quickly reflect in their standings. Additionally, while prompts and votes from all modes (battle, side-by-side, direct chat) are collected for transparent research, only anonymous battle votes contribute to the leaderboards, ensuring fairness. LM Arena also supports open research beyond the leaderboard, actively developing new artifacts and statistical methods to understand human preference with clarity and precision, including decomposing preference into components like tone, helpfulness, formatting, and emotional resonance. This research directly contributes to understanding how AI is perceived and trusted, which are key variables in understanding and mitigating hallucinations.

\begin{figure}[H]
    \centering
    \begin{minipage}{0.47\textwidth}
        \centering
        \includegraphics[width=\textwidth]{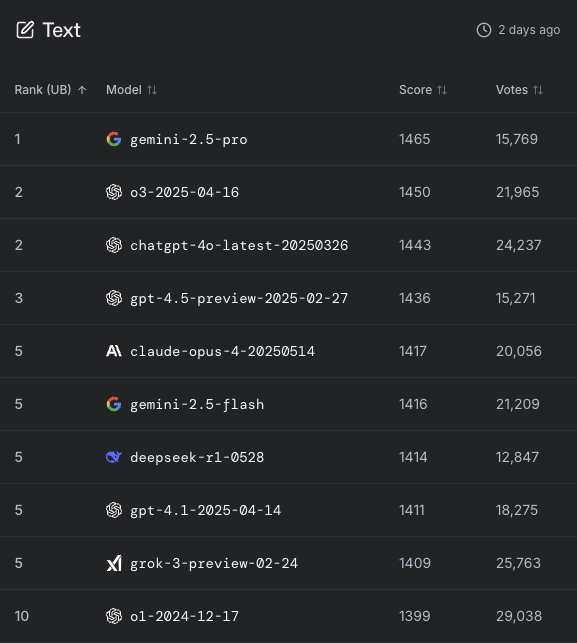}
        \caption{Sample visualization of models performance on text generation, retrieved on 9 July 2025}
        \label{fig:LMA_text}
    \end{minipage}\hfill
    \begin{minipage}{0.46\textwidth}
        \centering
        \includegraphics[width=\textwidth]{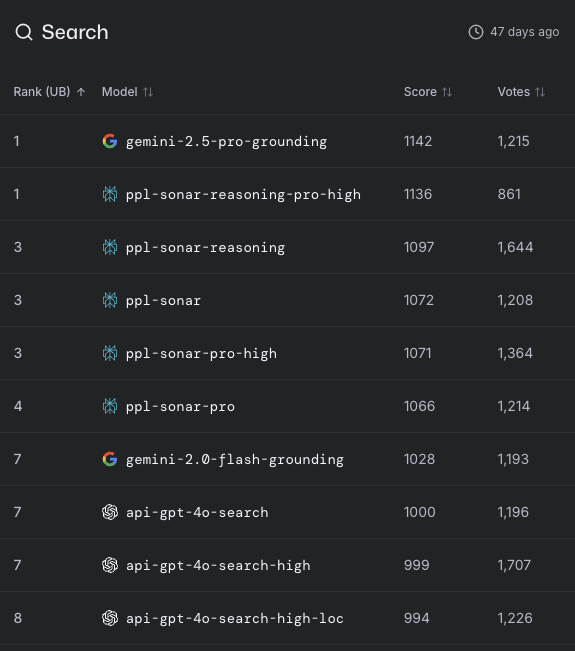}
        \caption{Sample visualization of models performance on web search for real-time information, external knowledge, and grounded citations, retrieved on 9 July 2025}
        \label{fig:LMA_search}
    \end{minipage}
    
    \label{fig:combined_LMA}
\end{figure}

\begin{figure}[H]
    \centering
    \begin{minipage}{0.48\textwidth}
        \centering
        \includegraphics[width=\textwidth]{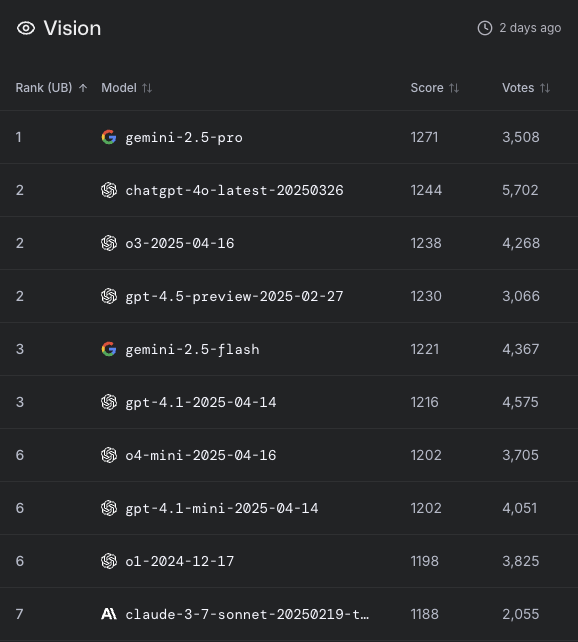}
        \caption{Sample visualization of models performance on generative AI models capable of understanding and processing visual inputs, retrieved on 9 July 2025}
        \label{fig:LMA_vision}
    \end{minipage}\hfill
    \begin{minipage}{0.47\textwidth}
        \centering
        \includegraphics[width=\textwidth]{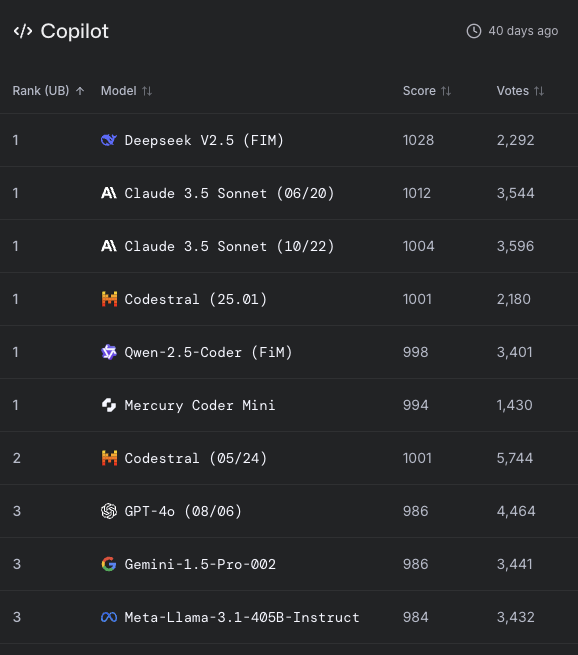}
        \caption{Sample visualization of models performance on how well AI coding assistants understand and generate code across various programming languages and tasks, retrieved on 9 July 2025}
        \label{fig:LMA_copilot_code}
    \end{minipage}
    
    \label{fig:combined_LMA_vision_code}
\end{figure}

Although not focused exclusively on hallucination, LM Arena provides real-world insights into which models are perceived as most helpful and trustworthy by end users. This user perception is a strong indicator of a model's ability to avoid hallucinations, as models that frequently hallucinate are unlikely to be rated as helpful or trustworthy. Therefore, LM Arena complements more technical hallucination metrics by offering a crucial perspective on the practical impact of hallucination on user experience and satisfaction. It underscores that reducing hallucinations is not just about factual correctness, but also about building user trust and providing genuinely useful AI assistance in real-world scenarios.

\vspace{0.5cm}
{\color{gray}\hrule}
\begin{center}
\section{Conclusions}
\bigskip
\end{center}
{\color{gray}\hrule}
\vspace{0.5cm}

\subsection{The complex nature and inevitable presence of LLM hallucinations}

Hallucination remains a pervasive and multifaceted challenge for LLMs, characterized by the generation of content that is plausible but factually incorrect, inconsistent, or entirely fabricated\cite{huang2025survey, elchafei2025span, joshi2025mitigating, lee2025hallucination}. As thoroughly evidenced throughout this report, these errors manifest in diverse forms, from core distinctions like intrinsic versus extrinsic hallucinations (contradicting input context versus inconsistency with training data or reality)\cite{bang2025hallulens, orgad2024llms, elchafei2025span, rafi2025reducing} and factuality versus faithfulness hallucinations (absolute correctness versus adherence to input)\cite{cao2021hallucinated, li2024dawn, maynez2020faithfulness, vishwanath2024faithfulness, malin2025review}, to specific manifestations such as factual errors, contextual and logical inconsistencies, temporal disorientation, ethical violations, and task-specific errors in domains like code generation and multimodal contexts\cite{joshi2025mitigating, chandler2024detecting, azaria2023internal, lee2025hallucination, cheong2024not,snyder2024early, liu2024exploring, agarwal2024codemirage, wu2025combating,  li2025detecting, elchafei2025span, xu2024hallucination, huang2025survey}.

These diverse issues stem from a complex interplay of factors, including data quality and biases\cite{joshi2025mitigating, gautam975impact, dai2024bias}, the inherent auto-regressive nature of LLM architectures\cite{li2024dawn, huang2025survey}, limitations in their training processes (such as exposure bias and capability misalignment)\cite{pozzi2025mitigating, lu2025maximum}, and stochastic decoding strategies\cite{huang2025survey}. Critically, formal theoretical proofs presented in this report indicate that hallucination is an innate and inevitable limitation for computable LLMs\cite{xu2024hallucination}, suggesting that complete elimination may be impossible regardless of architectural advancements or training refinements.

\subsection{Implications for detection, mitigation, and human interaction}

This profound implication necessitates a strategic shift from attempting complete eradication to developing robust detection mechanisms, implementing effective mitigation strategies, and ensuring continuous human oversight. The nuanced and task-specific nature of hallucinations underscores the need for granular approaches to their understanding and management, as a "one-size-fits-all" solution is unlikely to be effective.

The report has highlighted the critical role of cognitive and human factors in hallucination perception, emphasizing that user interfaces and interaction designs must incorporate strategies like calibrated uncertainty displays and source-grounding indicators to improve user resilience and trust\cite{yoo2024much, zeng2024justilm, lin2022teaching}. Furthermore, a comprehensive survey of evaluation benchmarks and metrics has revealed the ongoing challenges in standardized assessment, underscoring the need for unified, taxonomy-aware frameworks that can provide granular, diagnostic insights into hallucination types.

\subsection{Future directions for responsible LLM deployment}

Finally, the discussion of architectural and systemic mitigation strategies, including Toolformer-style augmentation\cite{schick2023toolformer} and RAG,\cite{joshi2025mitigating} alongside the introduction to web-based resources for monitoring LLM performance, provides practical directions for future development and deployment. This holistic understanding of hallucination types, their underlying causes, human interaction factors, evaluation methodologies, and mitigation techniques is paramount for developing more reliable, trustworthy, and safely deployable LLMs, particularly in high-stakes domains such as medicine and law, where the consequences of false information can be severe\cite{li2025detecting, lee2025hallucination, jiao2024navigating, gan2024navigating}. Continued research and responsible deployment practices, with a focus on human-in-the-loop validation and external safeguards, are essential for navigating the inherent limitations of LLMs and maximizing their transformative potential.

\vspace{0.5cm}
\bibliographystyle{plain}
\bibliography{references}

\begin{thebibliography}{100}

\bibitem{aboulela2025exploring}
Samar AboulEla, Paria Zabihitari, Nourhan Ibrahim, Majid Afshar, and Rasha Kashef.
\newblock Exploring rag solutions to reduce hallucinations in llms.
\newblock In {\em 2025 IEEE International systems Conference (SysCon)}, pages 1--8. IEEE, 2025.

\bibitem{agarwal2024codemirage}
Vibhor Agarwal, Yulong Pei, Salwa Alamir, and Xiaomo Liu.
\newblock Codemirage: Hallucinations in code generated by large language models.
\newblock {\em arXiv preprint arXiv:2408.08333}, 2024.

\bibitem{albrightimproving}
Jack Albright and Sheden Andemicael.
\newblock Improving llm mathematical reasoning capabilities using external tools.
\newblock {\em Stanford CS224R}, 2025.

\bibitem{an2024make}
Shengnan An, Zexiong Ma, Zeqi Lin, Nanning Zheng, Jian-Guang Lou, and Weizhu Chen.
\newblock Make your llm fully utilize the context.
\newblock {\em Advances in Neural Information Processing Systems}, 37:62160--62188, 2024.

\bibitem{andriushchenko2024agentharm}
Maksym Andriushchenko, Alexandra Souly, Mateusz Dziemian, Derek Duenas, Maxwell Lin, Justin Wang, Dan Hendrycks, Andy Zou, Zico Kolter, Matt Fredrikson, et~al.
\newblock Agentharm: A benchmark for measuring harmfulness of llm agents.
\newblock {\em arXiv preprint arXiv:2410.09024}, 2024.

\bibitem{azaria2023internal}
Amos Azaria and Tom Mitchell.
\newblock The internal state of an llm knows when it's lying.
\newblock {\em arXiv preprint arXiv:2304.13734}, 2023.

\bibitem{bang2025hallulens}
Yejin Bang, Ziwei Ji, Alan Schelten, Anthony Hartshorn, Tara Fowler, Cheng Zhang, Nicola Cancedda, and Pascale Fung.
\newblock Hallulens: Llm hallucination benchmark.
\newblock {\em arXiv preprint arXiv:2504.17550}, 2025.

\bibitem{beyer2021incoherence}
Anne Beyer, Sharid Lo{\'a}iciga, and David Schlangen.
\newblock Is incoherence surprising? targeted evaluation of coherence prediction from language models.
\newblock {\em arXiv preprint arXiv:2105.03495}, 2021.

\bibitem{binns2018s}
Reuben Binns, Max Van~Kleek, Michael Veale, Ulrik Lyngs, Jun Zhao, and Nigel Shadbolt.
\newblock 'it's reducing a human being to a percentage' perceptions of justice in algorithmic decisions.
\newblock In {\em Proceedings of the 2018 Chi conference on human factors in computing systems}, pages 1--14, 2018.

\bibitem{bn2025fact}
Suhas BN, Han-Chin Shing, Lei Xu, Mitch Strong, Jon Burnsky, Jessica Ofor, Jordan~R Mason, Susan Chen, Sundararajan Srinivasan, Chaitanya Shivade, et~al.
\newblock Fact-controlled diagnosis of hallucinations in medical text summarization.
\newblock {\em arXiv preprint arXiv:2506.00448}, 2025.

\bibitem{bubeck2023sparks}
S{\'e}bastien Bubeck, Varun Chadrasekaran, Ronen Eldan, Johannes Gehrke, Eric Horvitz, Ece Kamar, Peter Lee, Yin~Tat Lee, Yuanzhi Li, Scott Lundberg, et~al.
\newblock Sparks of artificial general intelligence: Early experiments with gpt-4, 2023.

\bibitem{byun2024reference}
Courtni Byun, Piper Vasicek, and Kevin Seppi.
\newblock This reference does not exist: an exploration of llm citation accuracy and relevance.
\newblock In {\em Proceedings of the Third Workshop on Bridging Human--Computer Interaction and Natural Language Processing}, pages 28--39, 2024.

\bibitem{cao2021hallucinated}
Meng Cao, Yue Dong, and Jackie Chi~Kit Cheung.
\newblock Hallucinated but factual! inspecting the factuality of hallucinations in abstractive summarization.
\newblock {\em arXiv preprint arXiv:2109.09784}, 2021.

\bibitem{chandler2024detecting}
Alex Chandler, Devesh Surve, and Hui Su.
\newblock Detecting errors through ensembling prompts (deep): an end-to-end llm framework for detecting factual errors.
\newblock {\em arXiv preprint arXiv:2406.13009}, 2024.

\bibitem{chang2022softmax}
Haw-Shiuan Chang and Andrew McCallum.
\newblock Softmax bottleneck makes language models unable to represent multi-mode word distributions.
\newblock In {\em Proceedings of the 60th Annual Meeting of the Association for Computational Linguistics}, volume~1, 2022.

\bibitem{chen2022towards}
Xiuying Chen, Mingzhe Li, Xin Gao, and Xiangliang Zhang.
\newblock Towards improving faithfulness in abstractive summarization.
\newblock {\em Advances in Neural Information Processing Systems}, 35:24516--24528, 2022.

\bibitem{chenghaozhu2025your}
ChenghaoZhu ChenghaoZhu, Nuo Chen, Yufei Gao, Yunyi Zhang, Prayag Tiwari, and Benyou Wang.
\newblock Is your llm outdated? a deep look at temporal generalization.
\newblock In {\em Proceedings of the 2025 Conference of the Nations of the Americas Chapter of the Association for Computational Linguistics: Human Language Technologies (Volume 1: Long Papers)}, pages 7433--7457, 2025.

\bibitem{cheong2022envisioning}
Inyoung Cheong, Aylin Caliskan, and Tadayoshi Kohno.
\newblock Envisioning legal mitigations for llm-based intentional and unintentional harms.
\newblock {\em Adm. Law J}, 2022.

\bibitem{cheong2024not}
Inyoung Cheong, King Xia, KJ~Kevin Feng, Quan~Ze Chen, and Amy~X Zhang.
\newblock (a) i am not a lawyer, but...: engaging legal experts towards responsible llm policies for legal advice.
\newblock In {\em Proceedings of the 2024 ACM Conference on Fairness, Accountability, and Transparency}, pages 2454--2469, 2024.

\bibitem{chhikara2025mind}
Prateek Chhikara.
\newblock Mind the confidence gap: Overconfidence, calibration, and distractor effects in large language models.
\newblock {\em arXiv preprint arXiv:2502.11028}, 2025.

\bibitem{cummings2017automation}
Mary~L Cummings.
\newblock Automation bias in intelligent time critical decision support systems.
\newblock In {\em Decision making in aviation}, pages 289--294. Routledge, 2017.

\bibitem{dai2024bias}
Sunhao Dai, Chen Xu, Shicheng Xu, Liang Pang, Zhenhua Dong, and Jun Xu.
\newblock Bias and unfairness in information retrieval systems: New challenges in the llm era.
\newblock In {\em Proceedings of the 30th ACM SIGKDD Conference on Knowledge Discovery and Data Mining}, pages 6437--6447, 2024.

\bibitem{dong2024building}
Yi~Dong, Ronghui Mu, Gaojie Jin, Yi~Qi, Jinwei Hu, Xingyu Zhao, Jie Meng, Wenjie Ruan, and Xiaowei Huang.
\newblock Building guardrails for large language models.
\newblock {\em arXiv preprint arXiv:2402.01822}, 2024.

\bibitem{doshi2017towards}
Finale Doshi-Velez and Been Kim.
\newblock Towards a rigorous science of interpretable machine learning.
\newblock {\em arXiv preprint arXiv:1702.08608}, 2017.

\bibitem{dzindolet2003role}
Mary~T Dzindolet, Scott~A Peterson, Regina~A Pomranky, Linda~G Pierce, and Hall~P Beck.
\newblock The role of trust in automation reliance.
\newblock {\em International journal of human-computer studies}, 58(6):697--718, 2003.

\bibitem{dziri2022origin}
Nouha Dziri, Sivan Milton, Mo~Yu, Osmar Zaiane, and Siva Reddy.
\newblock On the origin of hallucinations in conversational models: Is it the datasets or the models?
\newblock {\em arXiv preprint arXiv:2204.07931}, 2022.

\bibitem{elchafei2025span}
Passant Elchafei and Mervet Abu-Elkheir.
\newblock Span-level hallucination detection for llm-generated answers.
\newblock {\em arXiv preprint arXiv:2504.18639}, 2025.

\bibitem{es2024ragas}
Shahul Es, Jithin James, Luis~Espinosa Anke, and Steven Schockaert.
\newblock Ragas: Automated evaluation of retrieval augmented generation.
\newblock In {\em Proceedings of the 18th Conference of the European Chapter of the Association for Computational Linguistics: System Demonstrations}, pages 150--158, 2024.

\bibitem{favero2024enhancing}
Lucile Favero, Juan~Antonio P{\'e}rez-Ortiz, Tanja K{\"a}ser, and Nuria Oliver.
\newblock Enhancing critical thinking in education by means of a socratic chatbot.
\newblock {\em arXiv preprint arXiv:2409.05511}, 2024.

\bibitem{feng2024don}
Shangbin Feng, Weijia Shi, Yike Wang, Wenxuan Ding, Vidhisha Balachandran, and Yulia Tsvetkov.
\newblock Don't hallucinate, abstain: Identifying llm knowledge gaps via multi-llm collaboration.
\newblock {\em arXiv preprint arXiv:2402.00367}, 2024.

\bibitem{gan2024navigating}
Yuyou Gan, Yong Yang, Zhe Ma, Ping He, Rui Zeng, Yiming Wang, Qingming Li, Chunyi Zhou, Songze Li, Ting Wang, et~al.
\newblock Navigating the risks: A survey of security, privacy, and ethics threats in llm-based agents.
\newblock {\em arXiv preprint arXiv:2411.09523}, 2024.

\bibitem{gautam975impact}
Ankush~Ramprakash Gautam.
\newblock Impact of high data quality on llm hallucinations.
\newblock {\em International Journal of Computer Applications}, 975:8887, 2025.

\bibitem{gekhman2023trueteacher}
Zorik Gekhman, Jonathan Herzig, Roee Aharoni, Chen Elkind, and Idan Szpektor.
\newblock Trueteacher: Learning factual consistency evaluation with large language models.
\newblock {\em arXiv preprint arXiv:2305.11171}, 2023.

\bibitem{ghosh2024logical}
Bishwamittra Ghosh, Sarah Hasan, Naheed~Anjum Arafat, and Arijit Khan.
\newblock Logical consistency of large language models in fact-checking.
\newblock {\em arXiv preprint arXiv:2412.16100}, 2024.

\bibitem{holstein2019improving}
Kenneth Holstein, Jennifer Wortman~Vaughan, Hal Daum{\'e}~III, Miro Dudik, and Hanna Wallach.
\newblock Improving fairness in machine learning systems: What do industry practitioners need?
\newblock In {\em Proceedings of the 2019 CHI conference on human factors in computing systems}, pages 1--16, 2019.

\bibitem{honovich2022true}
Or~Honovich, Roee Aharoni, Jonathan Herzig, Hagai Taitelbaum, Doron Kukliansy, Vered Cohen, Thomas Scialom, Idan Szpektor, Avinatan Hassidim, and Yossi Matias.
\newblock True: Re-evaluating factual consistency evaluation.
\newblock {\em arXiv preprint arXiv:2204.04991}, 2022.

\bibitem{honovich2021q}
Or~Honovich, Leshem Choshen, Roee Aharoni, Ella Neeman, Idan Szpektor, and Omri Abend.
\newblock q$^2$: Evaluating factual consistency in knowledge-grounded dialogues via question generation and question answering.
\newblock {\em arXiv preprint arXiv:2104.08202}, 2021.

\bibitem{huang2025survey}
Lei Huang, Weijiang Yu, Weitao Ma, Weihong Zhong, Zhangyin Feng, Haotian Wang, Qianglong Chen, Weihua Peng, Xiaocheng Feng, Bing Qin, et~al.
\newblock A survey on hallucination in large language models: Principles, taxonomy, challenges, and open questions.
\newblock {\em ACM Transactions on Information Systems}, 43(2):1--55, 2025.

\bibitem{izacard2020leveraging}
Gautier Izacard and Edouard Grave.
\newblock Leveraging passage retrieval with generative models for open domain question answering.
\newblock {\em arXiv preprint arXiv:2007.01282}, 2020.

\bibitem{jiao2024navigating}
Junfeng Jiao, Saleh Afroogh, Yiming Xu, and Connor Phillips.
\newblock Navigating llm ethics: Advancements, challenges, and future directions.
\newblock {\em arXiv preprint arXiv:2406.18841}, 2024.

\bibitem{jin2023can}
Zhijing Jin, Jiarui Liu, Zhiheng Lyu, Spencer Poff, Mrinmaya Sachan, Rada Mihalcea, Mona Diab, and Bernhard Sch{\"o}lkopf.
\newblock Can large language models infer causation from correlation?
\newblock {\em arXiv preprint arXiv:2306.05836}, 2023.

\bibitem{joshi2025mitigating}
Satyadhar Joshi.
\newblock Mitigating llm hallucinations: A comprehensive review of techniques and architectures.
\newblock {\em preprints.org}, 2025.

\bibitem{kirk2023understanding}
Robert Kirk, Ishita Mediratta, Christoforos Nalmpantis, Jelena Luketina, Eric Hambro, Edward Grefenstette, and Roberta Raileanu.
\newblock Understanding the effects of rlhf on llm generalisation and diversity.
\newblock {\em arXiv preprint arXiv:2310.06452}, 2023.

\bibitem{kryscinski2019evaluating}
Wojciech Kry{\'s}ci{\'n}ski, Bryan McCann, Caiming Xiong, and Richard Socher.
\newblock Evaluating the factual consistency of abstractive text summarization.
\newblock {\em arXiv preprint arXiv:1910.12840}, 2019.

\bibitem{laban2022summac}
Philippe Laban, Tobias Schnabel, Paul~N Bennett, and Marti~A Hearst.
\newblock Summac: Re-visiting nli-based models for inconsistency detection in summarization.
\newblock {\em Transactions of the Association for Computational Linguistics}, 10:163--177, 2022.

\bibitem{le2020adversarial}
Ronan Le~Bras, Swabha Swayamdipta, Chandra Bhagavatula, Rowan Zellers, Matthew Peters, Ashish Sabharwal, and Yejin Choi.
\newblock Adversarial filters of dataset biases.
\newblock In {\em International conference on machine learning}, pages 1078--1088. Pmlr, 2020.

\bibitem{lee2025hallucination}
Yunseo Lee, John~Youngeun Song, Dongsun Kim, Jindae Kim, Mijung Kim, and Jaechang Nam.
\newblock Hallucination by code generation llms: Taxonomy, benchmarks, mitigation, and challenges.
\newblock {\em arXiv preprint arXiv:2504.20799}, 2025.

\bibitem{lewis2020retrieval}
Patrick Lewis, Ethan Perez, Aleksandra Piktus, Fabio Petroni, Vladimir Karpukhin, Naman Goyal, Heinrich K{\"u}ttler, Mike Lewis, Wen-tau Yih, Tim Rockt{\"a}schel, et~al.
\newblock Retrieval-augmented generation for knowledge-intensive nlp tasks.
\newblock {\em Advances in neural information processing systems}, 33:9459--9474, 2020.

\bibitem{li2025drift}
Hao Li, Xiaogeng Liu, Hung-Chun Chiu, Dianqi Li, Ning Zhang, and Chaowei Xiao.
\newblock Drift: Dynamic rule-based defense with injection isolation for securing llm agents.
\newblock {\em arXiv preprint arXiv:2506.12104}, 2025.

\bibitem{li2024dawn}
Junyi Li, Jie Chen, Ruiyang Ren, Xiaoxue Cheng, Wayne~Xin Zhao, Jian-Yun Nie, and Ji-Rong Wen.
\newblock The dawn after the dark: An empirical study on factuality hallucination in large language models.
\newblock {\em arXiv preprint arXiv:2401.03205}, 2024.

\bibitem{li2025detecting}
Ningke Li, Yahui Song, Kailong Wang, Yuekang Li, Ling Shi, Yi~Liu, and Haoyu Wang.
\newblock Detecting llm fact-conflicting hallucinations enhanced by temporal-logic-based reasoning.
\newblock {\em arXiv preprint arXiv:2502.13416}, 2025.

\bibitem{lin2004rouge}
Chin-Yew Lin.
\newblock Rouge: A package for automatic evaluation of summaries.
\newblock In {\em Text summarization branches out}, pages 74--81, 2004.

\bibitem{lin2021truthfulqa}
Stephanie Lin, Jacob Hilton, and Owain Evans.
\newblock Truthfulqa: Measuring how models mimic human falsehoods.
\newblock {\em arXiv preprint arXiv:2109.07958}, 2021.

\bibitem{lin2022teaching}
Stephanie Lin, Jacob Hilton, and Owain Evans.
\newblock Teaching models to express their uncertainty in words.
\newblock {\em arXiv preprint arXiv:2205.14334}, 2022.

\bibitem{lin2023mitigating}
Yong Lin, Hangyu Lin, Wei Xiong, Shizhe Diao, Jianmeng Liu, Jipeng Zhang, Rui Pan, Haoxiang Wang, Wenbin Hu, Hanning Zhang, et~al.
\newblock Mitigating the alignment tax of rlhf.
\newblock {\em arXiv preprint arXiv:2309.06256}, 2023.

\bibitem{ling2025bias}
Lin Ling, Fazle Rabbi, Song Wang, and Jinqiu Yang.
\newblock Bias unveiled: Investigating social bias in llm-generated code.
\newblock In {\em Proceedings of the AAAI Conference on Artificial Intelligence}, volume~39, pages 27491--27499, 2025.

\bibitem{liu2024exploring}
Fang Liu, Yang Liu, Lin Shi, Houkun Huang, Ruifeng Wang, Zhen Yang, Li~Zhang, Zhongqi Li, and Yuchi Ma.
\newblock Exploring and evaluating hallucinations in llm-powered code generation.
\newblock {\em arXiv preprint arXiv:2404.00971}, 2024.

\bibitem{liu2023trustworthy}
Yang Liu, Yuanshun Yao, Jean-Francois Ton, Xiaoying Zhang, Ruocheng Guo, Hao Cheng, Yegor Klochkov, Muhammad~Faaiz Taufiq, and Hang Li.
\newblock Trustworthy llms: a survey and guideline for evaluating large language models' alignment.
\newblock {\em arXiv preprint arXiv:2308.05374}, 2023.

\bibitem{lu2025maximum}
Tingmingke Lu.
\newblock Maximum hallucination standards for domain-specific large language models.
\newblock {\em arXiv preprint arXiv:2503.05481}, 2025.

\bibitem{luger2016like}
Ewa Luger and Abigail Sellen.
\newblock " like having a really bad pa" the gulf between user expectation and experience of conversational agents.
\newblock In {\em Proceedings of the 2016 CHI conference on human factors in computing systems}, pages 5286--5297, 2016.

\bibitem{malin2025review}
Ben Malin, Tatiana Kalganova, and Nikolaos Boulgouris.
\newblock A review of faithfulness metrics for hallucination assessment in large language models.
\newblock {\em IEEE Journal of Selected Topics in Signal Processing}, 2025.

\bibitem{malkin2021coherence}
Nikolay Malkin, Zhen Wang, and Nebojsa Jojic.
\newblock Coherence boosting: When your pretrained language model is not paying enough attention.
\newblock {\em arXiv preprint arXiv:2110.08294}, 2021.

\bibitem{mao2019neuro}
Jiayuan Mao, Chuang Gan, Pushmeet Kohli, Joshua~B Tenenbaum, and Jiajun Wu.
\newblock The neuro-symbolic concept learner: Interpreting scenes, words, and sentences from natural supervision.
\newblock {\em arXiv preprint arXiv:1904.12584}, 2019.

\bibitem{maynez2020faithfulness}
Joshua Maynez, Shashi Narayan, Bernd Bohnet, and Ryan McDonald.
\newblock On faithfulness and factuality in abstractive summarization.
\newblock {\em arXiv preprint arXiv:2005.00661}, 2020.

\bibitem{mehta2025embracing}
Seysha Mehta and Neil Mehta.
\newblock Embracing the illusion of explanatory depth: a strategic framework for using iterative prompting for integrating large language models in healthcare education.
\newblock {\em Medical Teacher}, 47(2):208--211, 2025.

\bibitem{min2023factscore}
Sewon Min, Kalpesh Krishna, Xinxi Lyu, Mike Lewis, Wen-tau Yih, Pang~Wei Koh, Mohit Iyyer, Luke Zettlemoyer, and Hannaneh Hajishirzi.
\newblock Factscore: Fine-grained atomic evaluation of factual precision in long form text generation.
\newblock {\em arXiv preprint arXiv:2305.14251}, 2023.

\bibitem{nickerson1998confirmation}
Raymond~S Nickerson.
\newblock Confirmation bias: A ubiquitous phenomenon in many guises.
\newblock {\em Review of general psychology}, 2(2):175--220, 1998.

\bibitem{ohayon2000prevalence}
Maurice~M Ohayon.
\newblock Prevalence of hallucinations and their pathological associations in the general population.
\newblock {\em Psychiatry research}, 97(2-3):153--164, 2000.

\bibitem{omar2025benchmarking}
Mahmud Omar, Reem Agbareia, Benjamin~S Glicksberg, Girish~N Nadkarni, and Eyal Klang.
\newblock Benchmarking the confidence of large language models in answering clinical questions: cross-sectional evaluation study.
\newblock {\em JMIR Medical Informatics}, 13:e66917, 2025.

\bibitem{orgad2024llms}
Hadas Orgad, Michael Toker, Zorik Gekhman, Roi Reichart, Idan Szpektor, Hadas Kotek, and Yonatan Belinkov.
\newblock Llms know more than they show: On the intrinsic representation of llm hallucinations.
\newblock {\em arXiv preprint arXiv:2410.02707}, 2024.

\bibitem{o2025confirmation}
Daniel~E O’Leary.
\newblock Confirmation and specificity biases in large language models: An explorative study.
\newblock {\em IEEE Intelligent Systems}, 40(1):63--68, 2025.

\bibitem{padiu2024extent}
Bogdan Padiu, Radu Iacob, Traian Rebedea, and Mihai Dascalu.
\newblock To what extent have llms reshaped the legal domain so far? a scoping literature review.
\newblock {\em Information}, 15(11):662, 2024.

\bibitem{pal2023med}
Ankit Pal, Logesh~Kumar Umapathi, and Malaikannan Sankarasubbu.
\newblock Med-halt: Medical domain hallucination test for large language models.
\newblock {\em arXiv preprint arXiv:2307.15343}, 2023.

\bibitem{pandit2025medhallu}
Shrey Pandit, Jiawei Xu, Junyuan Hong, Zhangyang Wang, Tianlong Chen, Kaidi Xu, and Ying Ding.
\newblock Medhallu: A comprehensive benchmark for detecting medical hallucinations in large language models.
\newblock {\em arXiv preprint arXiv:2502.14302}, 2025.

\bibitem{papineni2002bleu}
Kishore Papineni, Salim Roukos, Todd Ward, and Wei-Jing Zhu.
\newblock Bleu: a method for automatic evaluation of machine translation.
\newblock In {\em Proceedings of the 40th annual meeting of the Association for Computational Linguistics}, pages 311--318, 2002.

\bibitem{parmar2024towards}
Mihir Parmar, Hanieh Deilamsalehy, Franck Dernoncourt, Seunghyun Yoon, Ryan~A Rossi, and Trung Bui.
\newblock Towards enhancing coherence in extractive summarization: Dataset and experiments with llms.
\newblock {\em arXiv preprint arXiv:2407.04855}, 2024.

\bibitem{petroni2020kilt}
Fabio Petroni, Aleksandra Piktus, Angela Fan, Patrick Lewis, Majid Yazdani, Nicola De~Cao, James Thorne, Yacine Jernite, Vladimir Karpukhin, Jean Maillard, et~al.
\newblock Kilt: a benchmark for knowledge intensive language tasks.
\newblock {\em arXiv preprint arXiv:2009.02252}, 2020.

\bibitem{pozzi2025mitigating}
Andrea Pozzi, Alessandro Incremona, Daniele Tessera, and Daniele Toti.
\newblock Mitigating exposure bias in large language model distillation: an imitation learning approach.
\newblock {\em Neural Computing and Applications}, pages 1--17, 2025.

\bibitem{rafi2025reducing}
Shaik Rafi, Lenin Laitonjam, and Ranjita Das.
\newblock Reducing extrinsic hallucination in multimodal abstractive summaries with post-processing technique.
\newblock {\em Neural Computing and Applications}, pages 1--21, 2025.

\bibitem{reber1999effects}
Rolf Reber and Norbert Schwarz.
\newblock Effects of perceptual fluency on judgments of truth.
\newblock {\em Consciousness and cognition}, 8(3):338--342, 1999.

\bibitem{ren2023investigating}
Ruiyang Ren, Yuhao Wang, Yingqi Qu, Wayne~Xin Zhao, Jing Liu, Hao Tian, Hua Wu, Ji-Rong Wen, and Haifeng Wang.
\newblock Investigating the factual knowledge boundary of large language models with retrieval augmentation.
\newblock {\em arXiv preprint arXiv:2307.11019}, 2023.

\bibitem{roustan2025clinicians}
Dimitri Roustan, Fran{\c{c}}ois Bastardot, et~al.
\newblock The clinicians’ guide to large language models: A general perspective with a focus on hallucinations.
\newblock {\em Interactive journal of medical research}, 14(1):e59823, 2025.

\bibitem{rozenblit2002misunderstood}
Leonid Rozenblit and Frank Keil.
\newblock The misunderstood limits of folk science: An illusion of explanatory depth.
\newblock {\em Cognitive science}, 26(5):521--562, 2002.

\bibitem{saad2023ares}
Jon Saad-Falcon, Omar Khattab, Christopher Potts, and Matei Zaharia.
\newblock Ares: An automated evaluation framework for retrieval-augmented generation systems.
\newblock {\em arXiv preprint arXiv:2311.09476}, 2023.

\bibitem{salemi2024evaluating}
Alireza Salemi and Hamed Zamani.
\newblock Evaluating retrieval quality in retrieval-augmented generation.
\newblock In {\em Proceedings of the 47th International ACM SIGIR Conference on Research and Development in Information Retrieval}, pages 2395--2400, 2024.

\bibitem{schick2023toolformer}
Timo Schick, Jane Dwivedi-Yu, Roberto Dess{\`\i}, Roberta Raileanu, Maria Lomeli, Eric Hambro, Luke Zettlemoyer, Nicola Cancedda, and Thomas Scialom.
\newblock Toolformer: Language models can teach themselves to use tools.
\newblock {\em Advances in Neural Information Processing Systems}, 36:68539--68551, 2023.

\bibitem{scialom2021questeval}
Thomas Scialom, Paul-Alexis Dray, Patrick Gallinari, Sylvain Lamprier, Benjamin Piwowarski, Jacopo Staiano, and Alex Wang.
\newblock Questeval: Summarization asks for fact-based evaluation.
\newblock {\em arXiv preprint arXiv:2103.12693}, 2021.

\bibitem{seth2024hallucinogen}
Ashish Seth, Dinesh Manocha, and Chirag Agarwal.
\newblock Hallucinogen: A benchmark for evaluating object hallucination in large visual-language models.
\newblock {\em arXiv preprint arXiv:2412.20622}, 2024.

\bibitem{shinn2023reflexion}
Noah Shinn, Federico Cassano, Ashwin Gopinath, Karthik Narasimhan, and Shunyu Yao.
\newblock Reflexion: Language agents with verbal reinforcement learning.
\newblock {\em Advances in Neural Information Processing Systems}, 36:8634--8652, 2023.

\bibitem{shuster2021retrieval}
Kurt Shuster, Spencer Poff, Moya Chen, Douwe Kiela, and Jason Weston.
\newblock Retrieval augmentation reduces hallucination in conversation.
\newblock {\em arXiv preprint arXiv:2104.07567}, 2021.

\bibitem{singhal2023large}
Karan Singhal, Shekoofeh Azizi, Tao Tu, S~Sara Mahdavi, Jason Wei, Hyung~Won Chung, Nathan Scales, Ajay Tanwani, Heather Cole-Lewis, Stephen Pfohl, et~al.
\newblock Large language models encode clinical knowledge.
\newblock {\em Nature}, 620(7972):172--180, 2023.

\bibitem{snyder2024early}
Ben Snyder, Marius Moisescu, and Muhammad~Bilal Zafar.
\newblock On early detection of hallucinations in factual question answering.
\newblock In {\em Proceedings of the 30th ACM SIGKDD Conference on Knowledge Discovery and Data Mining}, pages 2721--2732, 2024.

\bibitem{taubenfeld2024systematic}
Amir Taubenfeld, Yaniv Dover, Roi Reichart, and Ariel Goldstein.
\newblock Systematic biases in llm simulations of debates.
\newblock {\em arXiv preprint arXiv:2402.04049}, 2024.

\bibitem{tian2025codehalu}
Yuchen Tian, Weixiang Yan, Qian Yang, Xuandong Zhao, Qian Chen, Wen Wang, Ziyang Luo, Lei Ma, and Dawn Song.
\newblock Codehalu: Investigating code hallucinations in llms via execution-based verification.
\newblock In {\em Proceedings of the AAAI Conference on Artificial Intelligence}, volume~39, pages 25300--25308, 2025.

\bibitem{uceda2024reasoning}
Rosario Uceda-Sosa, Karthikeyan~Natesan Ramamurthy, Maria Chang, and Moninder Singh.
\newblock Reasoning about concepts with llms: Inconsistencies abound.
\newblock {\em arXiv preprint arXiv:2405.20163}, 2024.

\bibitem{vishwanath2024faithfulness}
Prathiksha~Rumale Vishwanath, Simran Tiwari, Tejas~Ganesh Naik, Sahil Gupta, Dung~Ngoc Thai, Wenlong Zhao, SUNJAE KWON, Victor Ardulov, Karim Tarabishy, Andrew McCallum, et~al.
\newblock Faithfulness hallucination detection in healthcare ai.
\newblock In {\em Artificial Intelligence and Data Science for Healthcare: Bridging Data-Centric AI and People-Centric Healthcare}, 2024.

\bibitem{wang2024videohallucer}
Yuxuan Wang, Yueqian Wang, Dongyan Zhao, Cihang Xie, and Zilong Zheng.
\newblock Videohallucer: Evaluating intrinsic and extrinsic hallucinations in large video-language models.
\newblock {\em arXiv preprint arXiv:2406.16338}, 2024.

\bibitem{wu2025combating}
Shengqiong Wu, Hao Fei, Liangming Pan, William~Yang Wang, Shuicheng Yan, and Tat-Seng Chua.
\newblock Combating multimodal llm hallucination via bottom-up holistic reasoning.
\newblock In {\em Proceedings of the AAAI Conference on Artificial Intelligence}, volume~39, pages 8460--8468, 2025.

\bibitem{xu2023llm}
Xilie Xu, Keyi Kong, Ning Liu, Lizhen Cui, Di~Wang, Jingfeng Zhang, and Mohan Kankanhalli.
\newblock An llm can fool itself: A prompt-based adversarial attack.
\newblock {\em arXiv preprint arXiv:2310.13345}, 2023.

\bibitem{xu2024hallucination}
Ziwei Xu, Sanjay Jain, and Mohan Kankanhalli.
\newblock Hallucination is inevitable: An innate limitation of large language models.
\newblock {\em arXiv preprint arXiv:2401.11817}, 2024.

\bibitem{yin2023llm}
Wenpeng Yin, Qinyuan Ye, Pengfei Liu, Xiang Ren, and Hinrich Sch{\"u}tze.
\newblock Llm-driven instruction following: Progresses and concerns.
\newblock In {\em Proceedings of the 2023 Conference on Empirical Methods in Natural Language Processing: Tutorial Abstracts}, pages 19--25, 2023.

\bibitem{yoo2024much}
Minji Yoo.
\newblock How much should we trust llm-based measures for accounting and finance research?
\newblock {\em Available at SSRN}, 2024.

\bibitem{zeng2024justilm}
Fengzhu Zeng and Wei Gao.
\newblock Justilm: Few-shot justification generation for explainable fact-checking of real-world claims.
\newblock {\em Transactions of the Association for Computational Linguistics}, 12:334--354, 2024.

\bibitem{zhang2019bertscore}
Tianyi Zhang, Varsha Kishore, Felix Wu, Kilian~Q Weinberger, and Yoav Artzi.
\newblock Bertscore: Evaluating text generation with bert.
\newblock {\em arXiv preprint arXiv:1904.09675}, 2019.

\bibitem{zhang2024knowledge}
Yuji Zhang, Sha Li, Jiateng Liu, Pengfei Yu, Yi~R Fung, Jing Li, Manling Li, and Heng Ji.
\newblock Knowledge overshadowing causes amalgamated hallucination in large language models.
\newblock {\em arXiv preprint arXiv:2407.08039}, 2024.

\bibitem{zhang2025law}
Yuji Zhang, Sha Li, Cheng Qian, Jiateng Liu, Pengfei Yu, Chi Han, Yi~R Fung, Kathleen McKeown, Chengxiang Zhai, Manling Li, et~al.
\newblock The law of knowledge overshadowing: Towards understanding, predicting, and preventing llm hallucination.
\newblock {\em arXiv preprint arXiv:2502.16143}, 2025.

\bibitem{zhou2023making}
Xin Zhou, Yi~Lu, Ruotian Ma, Tao Gui, Qi~Zhang, and Xuanjing Huang.
\newblock Making harmful behaviors unlearnable for large language models.
\newblock {\em arXiv preprint arXiv:2311.02105}, 2023.

\bibitem{zou2023universal}
Andy Zou, Zifan Wang, Nicholas Carlini, Milad Nasr, J~Zico Kolter, and Matt Fredrikson.
\newblock Universal and transferable adversarial attacks on aligned language models.
\newblock {\em arXiv preprint arXiv:2307.15043}, 2023.

\bibitem{zuo2024medhallbench}
Kaiwen Zuo and Yirui Jiang.
\newblock Medhallbench: A new benchmark for assessing hallucination in medical large language models.
\newblock {\em arXiv preprint arXiv:2412.18947}, 2024.

\end{thebibliography}
\end{document}